\documentclass[journal,onecolumn]{IEEEtran}
\makeatletter
\def\ps@headings{%
	\def\@oddhead{\mbox{}\scriptsize\rightmark \hfil \thepage}%
	\def\@evenhead{\scriptsize\thepage \hfil \leftmark\mbox{}}%
	\def\@oddfoot{}%
	\def\@evenfoot{}}
\makeatother 

\makeatletter
\def\endthebibliography{%
	\def\@noitemerr{\@latex@warning{Empty `thebibliography' environment}}%
	\endlist
}
\makeatother

\usepackage[version=3]{acro}
\DeclareAcroProperty{alt2}
\DeclareAcroProperty{alt2-format}

\RenewAcroTemplate[list]{description}{%
  \acroheading
  \acropreamble
  \begin{description}
    \acronymsmapF{
      \item [%
          \acrowrite{short}%
          \acroifT{alt}{/}\acrowrite{alt}%
          \acroifT{alt2}{/}\acrowrite{alt2}%
        ]
        {\,\,\,\,\,\,\,\,\,\,\,}\acrowrite{long}
        \acroifanyT{foreign,extra}{ (}%
        \acroifT{foreign}{%
          \acrowrite{foreign}%
          \acroifT{foreign-short}{, \acrowrite{foreign-short}}%
          \acroifT{extra}{, }%
        }%
        \acrowrite{extra}%
        \acroifanyT{foreign,extra}{)}%
        \acropagefill
        \acropages
          {\acrotranslate{page}\nobreakspace}%
          {\acrotranslate{pages}\nobreakspace}%
    }
    { \item \AcroRerun }
  \end {description}
}

\DeclareAcronym{DRL}{
  short=DRL,
  long=Deep Reinforcement Learning,
}
\DeclareAcronym{RL}{
  short=RL,
  long=Reinforcement Learning,
}
\DeclareAcronym{DNN}{
  short=DNN,
  long=Deep Neural Network,
}
\DeclareAcronym{HDDQN}{
  short=HDDQN,
  long=Hierarchical Double Deep Q Network,
}
\DeclareAcronym{DQN}{
  short=DQN,
  long=Deep Q Network,
}
\DeclareAcronym{DQL}{
  short=DQL,
  long=Double Q-learning,
}
\DeclareAcronym{ML}{
  short=ML,
  long=Machine Learning,
}
\DeclareAcronym{DDQN}{
  short=DDQN,
  long=Double Deep Q Network,
}
\DeclareAcronym{TD-DDPG}{
  short=TD-DDPG,
  long=Twin Delayed-Deep Deterministic Policy Gradient,
}
\DeclareAcronym{HER}{
  short=HER,
  long=Hindsight Experience Replay,
}
\DeclareAcronym{HAC}{
  short=HAC,
  long=Hierarchical Actor-Critic,
}
\DeclareAcronym{SAC}{
  short=SAC,
  long=Soft Actor-Critic,
}
\DeclareAcronym{SMDP}{
  short=SMDP,
  long=Semi-Markov Decision Process,
}
\DeclareAcronym{RTEM}{
  short=RTEM,
  long=Real Time Energy Market,
}
\DeclareAcronym{SoC}{
  short=SoC,
  long=State of Charge,
}
\DeclareAcronym{V2G}{
  short=V2G,
  long=Vehicle-to-Grid,
}
\DeclareAcronym{GHG}{
  short=GHG,
  long=GreenHouse Gas,
}
\DeclareAcronym{RES}{
  short=RES,
  long=Renewable Energy Source,
}
\DeclareAcronym{RO}{
  short=RO,
  long=Robust Optimization,
}
\DeclareAcronym{IA}{
  short=IA,
  long=Immune Algorithm,
}
\DeclareAcronym{GA}{
  short=GA,
  long=Genetic Algorithm,
}
\DeclareAcronym{EBCSP}{ 
short = EBCSP, 
long = Electric Bus Charging Scheduling Problem,
short-format = \scshape,
}

\DeclareAcronym{EB}{
  short=EB,
  long=Electric Bus,
}
\DeclareAcronym{HRL}{
  short=HRL,
  long=Hierarchical Reinforcement Learning,
}
\DeclareAcronym{HDRL}{
  short=HDRL,
  long=Hierarchical Deep Reinforcement Learning,
}
\DeclareAcronym{MILP}{
  short=MILP,
  long=Mixed Integer Linear Programming,
}

\DeclareAcronym{MDP}{
  short=MDP,
  long=Markov Decision Process,
}
\DeclareAcronym{MRP}{
  short=MRP,
  long=Markov Reward Process,
}
\DeclareAcronym{TOU}{
  short=TOU,
  long=Time-of-Use,
}

\usepackage{graphicx}
\usepackage{amsmath, amsfonts,epsfig, multirow, floatflt}
\usepackage{amssymb}
\usepackage{subfigure}
\usepackage{cite}
\usepackage{algorithm}
\usepackage{algorithmicx}
\usepackage{algpseudocode}
\usepackage{hyperref}
\usepackage{enumitem}
\usepackage{diagbox}
\usepackage{mathrsfs}
\usepackage{url}
\usepackage{color}
\usepackage{xcolor}
\usepackage{colortbl}

\usepackage{caption}
\usepackage{hhline}
\usepackage{array}
\usepackage{booktabs}
\usepackage{amsthm}
\usepackage{bm}
\usepackage{setspace}

\begin{document}
	\title{Electric Bus Charging Schedules Relying on Real Data-Driven Targets Based on Hierarchical Deep Reinforcement Learning}
	

    \author{JIAJU QI$^{1}$, STUDENT MEMBER, IEEE, LEI LEI$^{1}$, SENIOR MEMBER, IEEE, THORSTEINN JONSSON$^{2}$, AND LAJOS HANZO$^{3}$, LIFE FELLOW, IEEE
    \thanks{$^{1}$J. Qi and L. Lei are with the School of Engineering, University of Guelph, Guelph, ON N1G 2W1, Canada, {\tt\small jiaju@uoguelph.ca; leil@uoguelph.ca} }
\thanks{$^{2}$T. Jonsson is with EthicalAI, Waterloo, ON N2L 0C7, Canada, {\tt\small thor@ethicalairesearch.com} }
\thanks{$^{3}$L. Hanzo is with the School of Electronics and Computer Science, University of Southampton, Southampton, SO17 1BJ, UK, {\tt\small hanzo@soton.ac.uk} }}
	

\maketitle

\begin{abstract}
The charging scheduling problem of Electric Buses (EBs) is investigated based on Deep Reinforcement Learning (DRL). A Markov Decision Process (MDP) is conceived, where the time horizon includes multiple charging and operating periods in a day, while each period is further divided into multiple time steps. To overcome the challenge of long-range multi-phase planning with sparse reward, we conceive Hierarchical DRL (HDRL) for decoupling the original MDP into a high-level Semi-MDP (SMDP) and multiple low-level MDPs. The Hierarchical Double Deep Q-Network (HDDQN)-Hindsight Experience Replay (HER) algorithm is proposed for simultaneously solving the decision problems arising at different temporal resolutions. As a result, the high-level agent learns an effective policy for prescribing the charging targets for every charging period, while the low-level agent learns an optimal policy for setting the charging power of every time step within a single charging period, with the aim of minimizing the charging costs while meeting the charging target. It is proved that the flat policy constructed by superimposing the optimal high-level policy and the optimal low-level policy performs as well as the optimal policy of the original MDP. Since jointly learning both levels of policies is challenging due to the non-stationarity of the high-level agent and the sampling inefficiency of the low-level agent, we divide the joint learning process into two phases and exploit our new HER algorithm to manipulate the experience replay buffers for both levels of agents. Numerical experiments are performed with the aid of real-world data to evaluate the performance of the proposed algorithm.  

\end{abstract}

\begin{IEEEkeywords}
Deep Reinforcement Learning; Electric Bus; Hierarchical Reinforcement learning; Charging Control
\end{IEEEkeywords}

{\color{black}\printacronyms[display=all]}

\section{Introduction}

\IEEEPARstart{I}{n} recent years, the adoption of Electric Buses (EBs) in public transportation fleets has been growing rapidly. Electrification of public transport reduces GreenHouse Gas (GHG) emissions, lowers maintenance and fuel costs, and provides the public with a more comfortable riding experience \cite{10415173}. With the increasing penetration of EBs, the question of how the charging cost can be reduced has become a key concern for bus companies \cite{wang2018bcharge}. Concurrently, in an attempt to spread the demand for electricity and efficiently utilize the Renewable Energy Sources (RES), power utility companies have adopted real-time electricity prices. Consequently, bus companies are presented with an effective measure to reduce costs and participate in the grid services by implementing intelligent charging schedules. \par

In order for EBs to fully take advantage of the time-varying electricity prices, intelligent strategies for charging scheduling are essential. Specifically, it is crucial to set an appropriate charging target, which is the target State of Charge (SoC) level at the end of a charging period. A high value may lead to unnecessary charging and high charging cost, while a low value could result in battery depletion during a trip.  \par


The optimization of EB charging schedules has been extensively explored, with existing contributions generally categorized into deterministic and stochastic schemes based on their system modeling approaches. Many studies, such as \cite{he2023battery, manzolli2022electric}, fall into the deterministic category. They consider system models where the parameter values are deterministic or constant, and typically derive optimal policies through solving Mixed Integer Linear Programming (MILP) problems. \par

While deterministic models do simplify the problem formulation and solution processes, they struggle to address two significant types of uncertainties encountered in practical scenarios, i.e., uncertainties in EB operation, such as random variations in travel time and energy consumption; plus uncertainties in the smart grid, including fluctuations in electricity prices. Consequently, system models that incorporate these uncertainties and stochastic elements offer a more realistic and applicable representation of the real world, ultimately leading to more reliable and efficient charging schedules. \par

However, the realistic consideration of uncertainties exacerbates the complexities of finding optimal charging targets. Firstly, since the amount of energy consumption during each trip depends on the time-varying traffic conditions along the bus route, the charging targets must be high enough to support the full trip taking into account these variations. For example, the charging targets have to be higher during rush hours, since it will take the EBs longer time and more energy to complete the trip. Secondly, the charging cost can be reduced by leveraging the real-time electricity prices. For example, EBs can create charging schedules that favor off-peak hours for charging, while discharging energy to the grid during peak hours in the Vehicle-to-Grid (V2G) mode \cite{10705088}. \par

In recent years, some studies within the stochastic category, such as \cite{liu2022optimal,bie2021optimization,tang2019robust,hu2022joint,zhou2022robust,liu2022robust}, have explored the uncertainties in EB travel time and energy consumption, utilizing methodologies such as Robust Optimization (RO) \cite{tang2019robust,hu2022joint,zhou2022robust,liu2022robust}, Immune Algorithms (IA) \cite{liu2022optimal}, and Genetic Algorithms (GA) \cite{bie2021optimization}. These approaches aim to enhance adaptability to the variable nature of EB operations. However, existing studies generally do not account for the uncertainties in real-time electricity prices, limiting their ability to optimize charging targets for different charging periods. Additionally, many existing investigations assume a constant charging power throughout the entire charging period, restricting EBs from discharging to the power grid via V2G and adjusting charging power in real time based on fluctuating electricity prices—both of which could further reduce charging costs.\par



To address the above limitations, this paper develops a charging strategy for EBs based on Deep Reinforcement Learning (DRL). DRL combines Reinforcement Learning (RL) with Deep Neural Networks (DNNs), and has shown significant potential to address uncertainties in operational environments in recent years. Compared with other techniques of dealing with uncertainty, such as RO, IA, and GA, DRL learns optimal policies directly from interactions with the environment, without relying on predefined stochastic models of the variables \cite{9277511}. Moreover, DRL dynamically learns and updates policies in real time, allowing it to swiftly adapt to changes in dynamic environments. Despite its great potential, there is currently a scarcity of literature applying DRL to charging scheduling for EBs \cite{9014160,yan2024mixed,WANG2024103516,10373906,zhou2023electric}, mainly due to the difficulty in designing an efficient and stable learning process.

Unlike existing DRL-based studies for EB charging, our work accounts for uncertainties in both electricity prices and traffic conditions while also incorporating flexible charging power decisions on a fine time-scale - such as every few minutes - to minimize the charging costs while ensuring sufficient battery energy for EB operations. Unfortunately, this task corresponds to a long-range multi-phase planning problem, which has not been addressed by the aforementioned DRL-based works. The solution to this problem is a key challenge in DRL, where the long action sequences give rise to issues such as slow convergence, high complexity, and insufficiently diverse exploration \cite{curtis2020flexible}. The situation further deteriorates when rewards are sparse, as in the EB charging problem when the penalty for running out of battery during trips occurs sporadically. As a result, learning from this undesirable outcome is challenging, because of its rare occurrence.\par


To overcome the aforementioned challenges, we design a novel Hierarchical DRL (HDRL) algorithm \cite{10.1145/3453160} in this paper. The main contributions are summarized as follows.


\begin{enumerate}

\item \emph{System Model}: Our system model extends beyond the bus transit system to include interactions with the electrical grid, embracing the uncertainties of electricity prices and integrating V2G capabilities. Furthermore,  charging scheduling decisions at a finer time-scale are considered, which determine the amount of real-time charging power, rather than relying on fixed charging power levels.

\item \emph{HDRL Model}: To address the challenges in solving the long-range multiple-phase planning problem with sparse reward, the charging target option is defined, which simultaneously serves as a temporally extended action and a closed-loop policy. By introducing options, the original MDP is decomposed into a novel two-level decision process including a high-level Semi-MDP (SMDP) and low-level MDPs. We demonstrate that effective policies can be learned from solving the high-level SMDP for prescribing data-driven charging targets. In order to learn a charging scheduling policy to realize the charging targets whenever possible with minimal charging cost, we define an intrinsic reward for the low-level MDPs. It is proved that the flat policy created by superimposing the optimal high-level policy and the optimal low-level policy performs as well as the optimal policy of the original MDP. Meanwhile, since the long-range effect is captured by the options, the low-level MDPs become isolated from each other, which can be trained over a shorter time horizon to improve the learning efficiency of long-range planning. 

\item  \emph{HDRL Solution}: A novel algorithm, namely the so-called Hierarchical Double Deep Q Network (HDDQN)-Hindsight Experience Replay (HER) is proposed for simultaneously solving the high-level SMDP and low-level MDPs while capturing the interplay between them. Specifically, both the high-level and low-level agents adopt the DDQN framework \cite{van2016deep}, but along with separate experience replay buffers. Both agents observe the system states and rewards from the environment for constructing their own rewards. The high-level agent prescribes specific charging target options that are fed to the low-level agent, which decides on the charging powers that are implemented in the environment. Finally, novel approaches based on HER are applied in order to reduce the impact of non-stationarity on the high-level agent and sample inefficiency on the low-level agent brought about by simultaneously learning multiple levels of policies. This innovation enables the proposed algorithm to learn more efficiently and converge faster.

\end{enumerate}

The rest of the paper is organized as follows. Section II critically appraises the state-of-the-art. The system model and MDP model are introduced in Sections III and IV, respectively, while the two-level decision process with options is formulated in Section V. The HDDQN-HER algorithm is proposed in Section VI. Finally, our experiments are highlighted in Section VII, and conclusions are offered in Section VIII.

\section{State-of-the-art}

\subsection{Research on EB charging scheduling problem}

There are three major charging technologies for EBs: (i) conductive charging; (ii) battery swapping; and (iii) wireless charging. The studies on conductive charging scheduling can be further divided into plug-in charging and pantograph charging. Both types are classified into two distinct approaches: (i) depot charging which adopts normal or slow chargers to charge EBs at bus depots overnight; (ii) opportunistic charging \footnote{This terminology is synonymous with the ``opportunity charging" in \cite{zhou2024charging}.} which utilizes fast chargers to charge EBs at terminal stations or stops \cite{zhou2024charging}. In the following, we will primarily review the existing literature on the EB Charging Scheduling Problem (EBCSP) for plug-in charging and opportunistic charging, which constitutes the main focus of this paper.

Broadly speaking, the EBCSP involves one or more EBs, a set of trips to be fulfilled by the EBs, and charging infrastructures. The objective is to optimize the charging schedule for minimizing costs, ensuring there is sufficient battery energy to support the EBs' trips, while adhering to various operational constraints, primarily concerning the bus schedules, power grid, and charger availability. \par


Traditionally, all the parameters in the EBCSP, such as the travel time and energy consumption for each trip, are assumed to be constant or known. Based on this deterministic information, the EBCSP is normally formulated as a MILP problem and solved by different methods such as Branch \& Price (BP) \cite{ji2023optimal}, the column generation based algorithm \cite{ZhangLe2021Oebf}, dynamic programming \cite{bao2023optimal}, and optimization solvers such as CPLIX \cite{manzolli2022electric,zhou2022electric,rinaldi2020mixed,he2023battery,he2020optimal}. 

There are different types of decisions on charging schedules. The simplest type is a binary variable indicating whether to charge or not, whenever an EB arrives at the terminal station after a trip. Most studies on binary charging decisions adopted a full charging policy, i.e., once an EB is assigned for charging, it must be fully charged before embarking on a new trip \cite{he2023battery, rinaldi2020mixed}. Given the constraints imposed by bus schedules, which may not provide ample time for full charging, the studies in \cite{whitaker2023network, ZhangLe2021Oebf,ji2023optimal} opted for partial charging, and derived the achieved SoC level of battery based on factors including the energy consumption rate and the length of time until the next departure of the EB. Given that a simplistic binary decision could result in unnecessarily high SoC level and increased charging cost, most studies on partial charging made decisions on the charging duration while assuming a fixed charging power \cite{bao2023optimal,xie2023collaborative,zhou2024electric}. In order to further enhance the flexibility in making charging scheduling decisions, \cite{he2020optimal} divided the charging period into smaller time intervals and optimized the charging power for each time interval. \par

However, MILP-based methods have several limitations. First, MILP becomes intractable for large-scale problems, with computation time growing exponentially. Additionally, its adaptability to dynamic environments is poor, as any environment change requires re-modeling and re-solving, further exacerbated by its high computational cost. Most importantly, in realistic scenarios, the travel time and energy consumption are inherently stochastic due to factors such as traffic flow and intersection waiting time. Therefore, their exact values are unknown when solving the EBCSP in practice. Since MILP relies on deterministic modeling, it struggles to effectively handle these uncertainties. To address this issue, He \emph{et al.} \cite{he2023battery} estimated an EB’s travel time and energy consumption using the K-means clustering algorithm, and plugged in the predicted values in the MILP model to derive the charging schedule. However, it is impossible to completely eliminate the prediction errors given current technology limitations. \par

In order to address the uncertainties associated with stochastic travel time and/or energy consumption, stochastic models have been formulated. These models inherently increase the complexity of the optimization algorithms designed for charging scheduling. For example, RO was widely used by the studies \cite{tang2019robust }, \cite{ hu2022joint }, \cite{ zhou2022robust}, and \cite{liu2022robust }, to achieve the dual goals of reducing charging costs and improving system robustness. Both \cite{tang2019robust} and \cite{hu2022joint} considered a fixed charging power, with the former making binary charging decisions, while the latter optimizing the charging time. Conversely, both \cite{zhou2022robust} and \cite{liu2022robust} optimized the charging power at smaller time-scales. Although RO can handle uncertainties and provide robust solutions within a predefined uncertainty set, it often leads to overly conservative policies, sacrificing performance to mitigate risks. RO requires modeling the uncertainty set and optimizing across multiple uncertain scenarios, resulting in high computational complexity, but its complexity is still lower than that of MILP. Additionally, RO lacks adaptability to dynamic environmental changes. 

Some studies opted for heuristic algorithms to address the uncertainties. For example, Liu \emph{et al.} \cite{liu2022optimal} employed the ant colony algorithm and IA to optimize charging plans under fixed and stochastic travel times, respectively, with a focus on charging time as the optimization variable. Bie \emph{et al.} \cite{bie2021optimization} presented the probability distribution function of trip energy consumption and adopted GA to reduce the charging costs as well as the delay in trip departure times. Compared to MILP, heuristic algorithms represented by GA have lower computational costs. However, they require multiple iterative optimizations, resulting in slow convergence. Additionally, GA typically operates offline and requires the entire algorithm to be re-run to integrate new information or accommodate changes in the environment, limiting its adaptability to dynamic environments. This limitation underscores the suitability of DRL for solving EBCSP. 

\begin{figure}[t]
    \centering
    \includegraphics[width=0.5\linewidth]{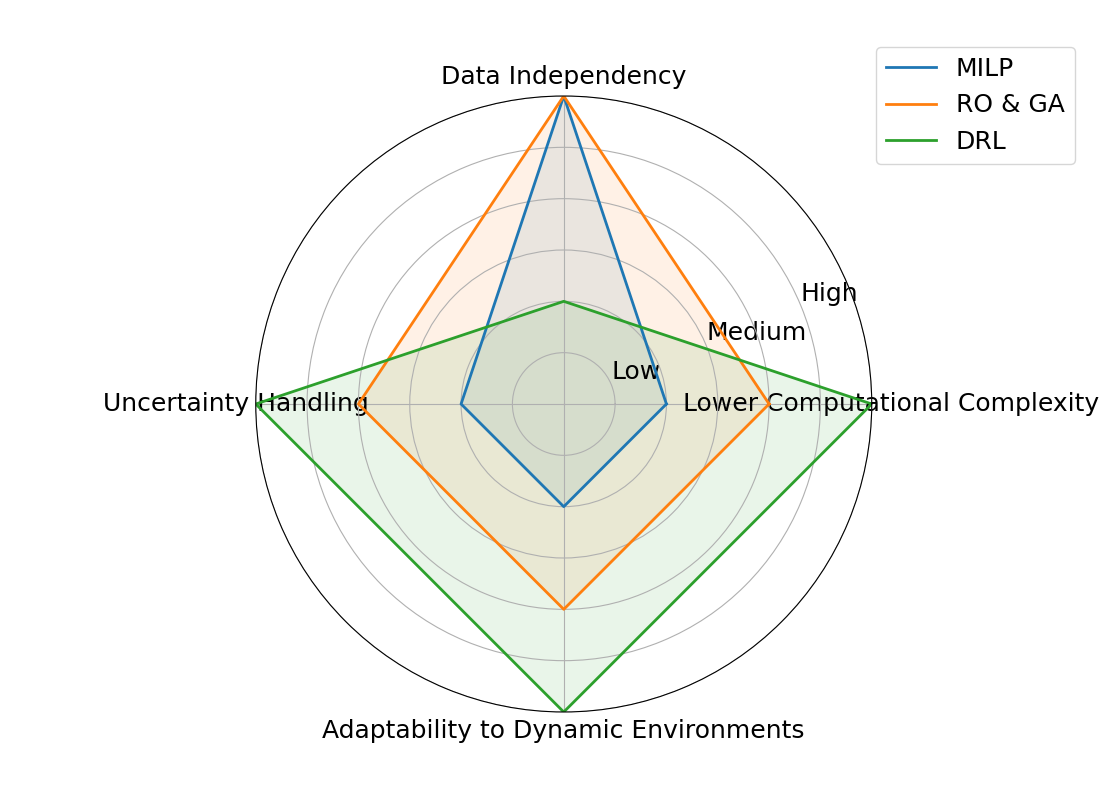}
    \caption{Comparison of MILP, RO, GA, and DRL in terms of four key criteria: lower computational complexity, data independency, uncertainty handling, and adaptability to dynamic environments.}
    \label{fig:radar-chart}
\end{figure}

Unlike traditional methods, DRL does not require precise modeling; instead, it learns directly from interactions with the environment, making it the most effective in handling uncertainty. It also demonstrates strong adaptability in dynamic environments. The training process of DRL requires a significant amount of time. However, once training is complete, DRL can make decisions quickly in real time during the deployment phase, as it only requires forward propagation through the trained neural network. However, a limitation of DRL is its heavy reliance on training data. Therefore, when designing new DRL algorithms, achieving higher sample efficiency is crucial. In Fig. \ref{fig:radar-chart}, we visually present the comparison of MILP, RO, GA, and DRL.\par 

There is relatively little research on DRL-based solutions for EBCSP. Among these studies, Wang \emph{et al.} \cite{WANG2024103516} combined clipped Double Q-learning (DQL) with Soft Actor-Critic (SAC) to simultaneously solve EB dispatching and charging scheduling problems, where the charging scheduling decision is a simple binary variable. Yan \emph{et al.} \cite{yan2024mixed} adopted Q-Learning and Twin Delayed Deep Deterministic Policy Gradient (TD-DDPG) to determine the EBs’ target SoC levels and their assigned service routes, while assuming fixed charging power. Chen \emph{et al.} \cite{9014160} used DQL to make decisions regarding the charging power for EBs upon their arrivals at the terminal station. However, the charging power remains constant for each charging period. By contrast, our work considers more flexible charging power decisions at a smaller time-scale. In summary, while prior works apply various classical DRL techniques, they face challenges due to the inherent long-range multi-phase planning problem when applied to our system model. Therefore, our study aims to address this complexity by introducing a hierarchical framework with options.


In addition to the parameters associated with the bus transit system, the characteristics of the smart grid have also been considered in some treatises. In order to reduce the charging cost, Manzolli \emph{et al.} \cite{manzolli2022electric} explored the potential of the V2G scheme, which allows EBs to sell electricity back to the grid. Meanwhile, some studies \cite{liu2022optimal,bao2023optimal,whitaker2023network,zhou2022robust} examined the impact of Time-of-Use (TOU) electricity tariffs, which facilitate learning of optimal charging policies that exploit time-varying electricity prices. However, the electricity prices are typically assumed to be known and static for different periods of the day in all the above studies. In recent years, with the increasing penetration of RES into the electric grid, the adoption of the Real-Time Energy Market (RTEM) \cite{RTEM} has been on the rise. This market allows participants to buy and sell wholesale electricity throughout the operating day, aiming to balance actual real-time demand with the fluctuating production of electricity \cite{ISOEng2024}. We direct interested readers to \cite{en11112974, LE2019258} for more information on how real-time electricity prices are determined and communicated to users. By transiting from static pricing contracts to real-time pricing contracts, the bus companies can potentially reduce their charging costs, while supporting the grid in achieving long-term stability in power supply \cite{fei2023exploring}. However, the inherent randomness of real-time electricity prices poses an additional challenge in solving EBCSP, where a sequence of charging schedule decisions must be made to minimize the charging cost without knowing the fluctuating electricity prices for the rest of the day. This challenge has been rarely touched upon in previous research. This knowledge gap is addressed in our paper. \par

Table \ref{related works} boldly contrasts our work to the existing literature on EBCSP in terms of key features. While most existing work aims for minimizing the charging cost and considers the more flexible partial charging option rather than full charging, only some of the studies take into account the uncertainties in energy consumption and/or travel time, and consider the charging power as an adjustable decision variable. In addition, existing treatises generally do not incorporate more advanced smart grid characteristics, such as the V2G mode and uncertainty in electricity price. Furthermore, the DRL-based algorithms are rarely adopted to solve the EBCSP, addressing the uncertainties in practical systems. As shown in Table \ref{related works}, our paper considers all the listed features, filling the corresponding gaps in this field. \par


\begin{table*}[t]
\centering
\color{black}
\small
\caption{Contrasting this paper to the literature on EBCSP.}
\begin{tabular}{|l|c|c|c|c|c|c|c|c|c|c|}
\hline
\textbf{Features}& \cite{he2023battery,rinaldi2020mixed}  & \begin{tabular}[l]{@{}l@{}}\cite{bao2023optimal,ji2023optimal,zhou2022electric,ZhangLe2021Oebf}  \\ \cite{xie2023collaborative,whitaker2023network,zhou2024electric}\end{tabular}   &  \cite{he2020optimal}&\cite{manzolli2022electric}&\cite{liu2022optimal,bie2021optimization,tang2019robust,hu2022joint}&\cite{liu2022robust,zhou2022robust}&\cite{yan2024mixed,WANG2024103516} & \cite{9014160} & Our work \\ \hline
\textbf{\textbf{\begin{tabular}[l]{@{}l@{}}Charging cost \\ minimization\end{tabular}}}   & \checkmark & \checkmark   & \checkmark   &\checkmark&\checkmark&\checkmark& &           &   \checkmark \\ \hline
\textbf{{\color{black}Partial charging}} &  &\checkmark&\checkmark&\checkmark &\checkmark&\checkmark & \checkmark &\checkmark &  \checkmark \\ \hline
\textbf{\textbf{\begin{tabular}[l]{@{}l@{}}Uncertainties in  \\ {\color{black}energy consumption} \\ {\color{black} and/or travel time}\end{tabular}}} &   &   &    &    &\checkmark&\checkmark& \checkmark& \checkmark &   \checkmark \\ \hline


\textbf{\textbf{\begin{tabular}[l]{@{}l@{}}Adjustable \\ charging power\end{tabular}}}&   & &\checkmark &  & & \checkmark&  & \checkmark  &  \checkmark  \\ \hline
\textbf{DRL-based algorithm}& &&&&&& \checkmark  &  \checkmark  &   \checkmark \\ \hline
\textbf{V2G mode} &  & & & \checkmark  & & & &  &  \checkmark \\ \hline
\textbf{\textbf{\begin{tabular}[l]{@{}l@{}}{\color{black}Uncertainty in} \\{\color{black}electricity price} \end{tabular}}} &  &                                                    &         &&&&          &                  &  \checkmark \\ \hline
\end{tabular}
\label{related works}
\end{table*}

\subsection{Hierarchical RL (HRL)}
The existing approaches on HRL are mainly developed based on three basic frameworks \cite{barto2003recent}, i.e., the option framework by Sutton \emph{et al.} \cite{sutton1999between}, MAXQ  by Dietterich \emph{et al.} \cite{dietterich2000hierarchical}, and the Hierarchy of Abstract Machines (HAMs) by Parr and Russell \cite{parr1997reinforcement}. In the most widely-used option framework, an option is associated with a subtask and can be regarded as a high-level action. It is represented by three key components including the initiation condition, the low-level intra-option policy that is used for choosing actions, and the termination probability function. It can be proved that any MDP having a fixed set of options is an SMDP, and its solution leads to a high-level policy over options that is used for prescribing the options. In contrast to MAXQ, the individual intra-option policies cannot be learned as an independent unit in the option framework. They should be integrated into the original MDP and optimized as a whole \cite{10.1145/3453160}. Thus, the optimality of learning a hierarchical policy is theoretically guaranteed. \par 

The studies on the option framework can be categorized into the approaches with or without subtask discovery. The approaches with subtask discovery, such as those in \cite{bacon2017option,klissarov2017learnings}, do not predefine the subtasks for the options. By contrast, the approaches without subtask discovery, such as the H-DQN by Kulkarni \emph{et al.} \cite{kulkarni2016hierarchical}, associate each option with a subgoal state for the low-level policy to achieve. In this case, some of the key components of option such as the termination function may not have to be learned, since an option may be deemed to be terminated, once the subgoal is achieved. In this paper, we adopt the option framework without subtask discovery and associate the options with the charging targets. In contrast to the existing HRL algorithms operating without subtask discovery \cite{kulkarni2016hierarchical,nachum2018data,levy2017learning}, we formally formulate our problem associated with the options, providing a unified framework for both option learning and subgoal learning.   \par

\begin{figure}[t]
\centering
\includegraphics[width=0.5\linewidth]{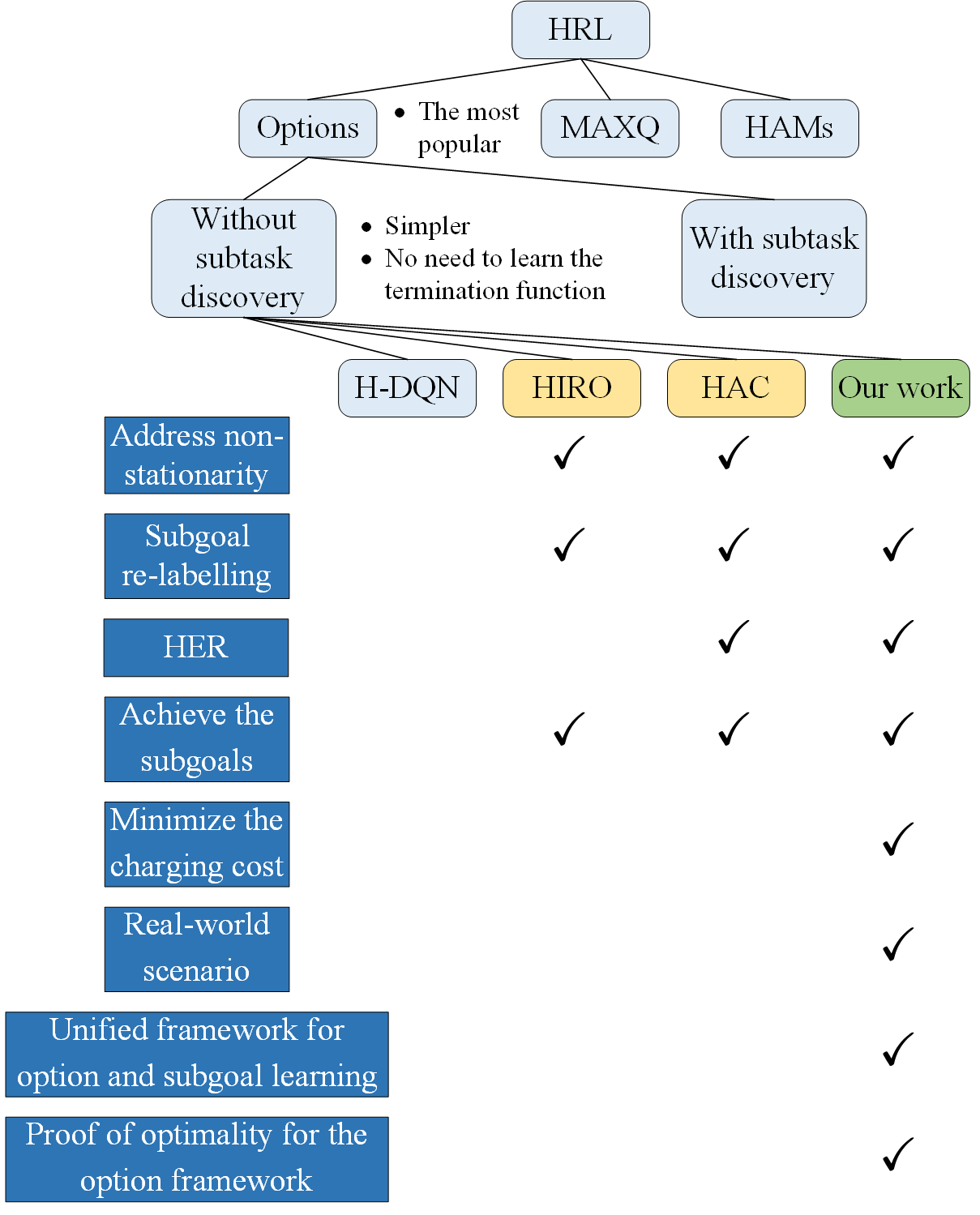}
\caption{The contributions of this paper in contrast to the literature on HRL.}
\label{related works HRL}
\end{figure}

An important challenge when simultaneously learning a hierarchy of policies is associated with the non-stationarity \cite{10.1145/3453160}, since the low-level policy may change during the training process, which leads to non-stationary reward functions and transition probabilities of the high-level SMDP. To address this challenge, the Hierarchical Reinforcement Learning with Off-policy Correction (HIRO) by Nachum \emph{et al.} \cite{nachum2018data} adopted a mechanism of subgoal re-labeling. The Hierarchical Actor-Critic (HAC) regime of Levy \emph{et al.} \cite{levy2017learning} integrates the ideas of Hindsight Experience Replay (HER) \cite{andrychowicz2017hindsight} with HIRO for further improvement. Since our objective in this paper involves not only achieving the charging target subgoals but also minimizing the charging cost, our problem is more challenging compared to the studies in \cite{nachum2018data} and \cite{levy2017learning}. Fig. \ref{related works HRL} shows the contributions of this paper in contrast to the literature of HRL. 

\section{System model}

The variables and parameters used in this paper are summarized in Table \ref{notations}. In general, we use cursive letters to represent a set, such as $\mathcal{S}$ and $\mathcal{A}$, uppercase letters to represent the total number of variables in a set, such as $T$ and $K$, and the corresponding lowercase letters to represent the variable indexes in a set, such as $t$ and $k$. In addition, following the convention of reinforcement learning, we generally use the capital letters with subscript $t$ to denote random variables, such as $S_t$ and $A_t$, while the corresponding lowercase letters represent sample values of the random variables, such as $s$ and $a$. We use uppercase letters to represent constant values, such as $E_{\max}$ and $E_{\min}$. Finally, we use lowercase letters to represent the functions, such as $r(S_t,A_t)$.\par
\begin{table*}[t]
\centering
\color{black}
\caption{Notation used in this paper}
\begin{tabular}[b]{p{2.5cm}<{\raggedright}p{14.0cm}<{\raggedright}}
\hline
\textbf{Notations}&\textbf{Description}\\
\hline
\specialrule{0em}{1pt}{1pt}
\multicolumn{2}{l}{\textbf{Sets}}  \\
\specialrule{0em}{1pt}{1pt}
$\mathcal{A}$ & The action space\\
$\mathcal{I}_\omega $ & The initiation set of states for option $\omega$\\
$\mathcal{S}^{+}$,$\mathcal{S}^{\rm T}$/$\mathcal{S}$ & The state space, the set of terminal/non-terminal states\\
$\varOmega$ & The option space\\
\specialrule{0em}{1pt}{1pt}
\multicolumn{2}{l}{\textbf{Parameters}}  \\
\specialrule{0em}{1pt}{1pt}
$B_t$ & The EB status at time step $t$, $0$ for operating periods and $1$ for charging periods \\
$C^\mathrm{end}$ &The penalty for EB to enter the terminal states\\
$C_{\max}$/$D_{\max}$ & The maximum absolute value of charging/discharging power of EBs\\
$E_t$ & The SoC level of the battery at time step $t$\\
$E_{\max}$/$E_{\min}$ & The maximum/minimum storage constraint of the battery\\
$H_t$ &  The historical electricity prices in the period spanning from time step $t-w_p$ up to time step $t$\\
$K$ & The number of operating periods in a day\\
$k$ & The index of an operating/charging period\\
$k_t$ & The index of the current charging/operating period at the time step $t$\\
$P_t$ & The electricity price at time step $t$\\
$S_t$ & The system state at time step $t$\\
$t_k$ & The random time step at which the EB arrives at the terminal station from the operating period $k$\\
$T_{k}^\mathrm{a}$ & The random number of time steps between two consecutive arrivals of the EB from operating periods $k$ and $k+1$ at $t_{k}$ and $t_{k+1}$, respectively\\
$T_{k}^\mathrm{d}$ & The fixed number of time steps between two consecutive departures of the EB for operating periods $k$ and $k+1$ according to the bus schedule \\
$T_k^{\rm c}$/$T_k^{\rm o}$ & The random number of time steps in the charging/operating period $k$\\
$w_p$ &The length of the time window to look into the past prices \\
$\Delta t$ & The duration of each time step\\
$\tau_t$ & The number of remaining time steps from time step $t$ to the next departure time\\
\specialrule{0em}{1pt}{1pt}
\multicolumn{2}{l}{\textbf{Decision Variables}}  \\
\specialrule{0em}{1pt}{1pt}
$A_t$ & The action at time step $t$\\
$C_t$ & The charging/discharging power of the battery at time step $t$\\
$\omega_t$ & The charging target option at time step $t$\\
\multicolumn{2}{l}{\textbf{Functions}}  \\
\specialrule{0em}{1pt}{1pt}
$c^{\mathrm{tar}}(S_t,\omega_t)$ & The penalty of not realizing the charging target \\
$r(S_t, A_t)$ & The reward function\\
$r^{\mathrm{H}}\left( S_t,\omega_t \right)$ & The expected cumulative reward within one transition of SMDP\\
$r^{\mathrm{L}}\left( S_t,\omega_t,A_t \right)$ & The  intrinsic reward of the low-level MDPs\\
$\beta_\omega(S_t)$ & The termination condition of option $\omega$\\
$\varGamma_{t}(B_t)$ & The probability that the current period is terminated at time step $t$\\
$\pi_\omega(S_t)$ & The intra-option policy for option $\omega$\\
$\mu(S_t)$ & The policy over options\\
\specialrule{0.05em}{2pt}{0pt}
\end{tabular}
\label{notations}
\end{table*}
In this paper, we consider an EB routinely traveling along a bus route that includes several bus stops and a terminal station. When the EB arrives at the terminal station, its battery pack can be charged. Therefore, the status of the EB can be divided into two alternating periods. One is the charging period, in which the EB stays at the terminal station for charging. The other is the operating period, in which the EB travels along the designated route. 

\begin{figure}[t]
\centering
\includegraphics[width=0.57\linewidth]{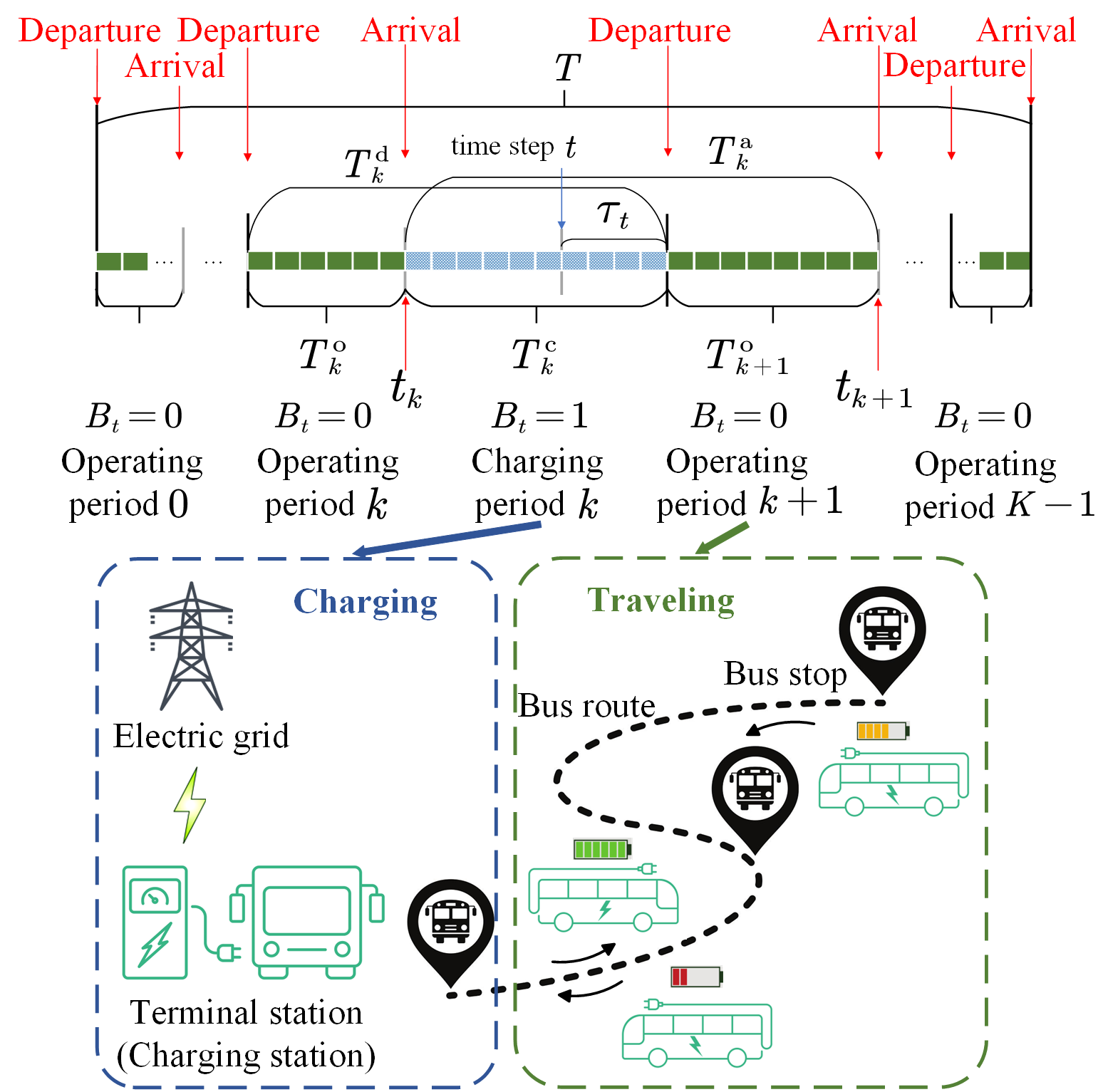}
\caption{The schematic diagram of the system model for the EB charging problem.}
\label{system_model}
\end{figure}

As shown in Fig. \ref{system_model}, consider that there are $K$ operating periods for the EB in a single day according to its bus schedule, where each operating period is indexed by $k\in\{0, \dots, K-1 \}$. Additionally, each operating period $k\in\{0, \dots, K-2 \}$ is followed by a charging period with the same index $k$. Given our focus on opportunistic charging, which involves fast chargers at terminal stations, we exclude depot charging that typically occurs overnight after the last operating period $K-1$ of the day. Therefore, the time horizon considered in this paper starts and ends with an operating period, and there are $K-1$ charging periods sandwiched between $K$ operating periods.  \par

The time in a full day is discretized into $T$ equal-length time steps, indexed by $\{0, \dots, T-1 \}$. The duration of each time step is denoted as $\Delta t$. Let $T_{k}^{\rm o}$, $\forall k\in\{0, \dots, K-1 \}$ and $T_{k}^{\rm c}$, $\forall k\in\{0, \dots, K-2 \}$ be random variables that represent the number of time steps in the operating period $k$ and the charging period $k$, respectively. Note that the variables $T_{k}^{\rm c}$ and $T_{k}^{\rm o}$ are random due to the uncertainty in the arrival times of the EBs at the terminal station, which is caused by the stochastic traffic conditions along the bus route. The random time step at which the EB arrives at the terminal station from the operating period $k$ is denoted by $t_{k}$, where $k\in\{0, \dots, K-1\}$. Then, the EB enters the charging period $k$ upon its arrival at $t_k$, where $k\in\{0, \dots, K-2\}$. \par

Let $T_{k}^\mathrm{d}$, $\forall k\in\{0, \dots, K-2 \}$ denote the fixed number of time steps between two consecutive departures of the EB for operating periods $k$ and $k+1$ according to the bus schedule. This corresponds to the sum duration of a pair of consecutive operating and charging periods, i.e., $T_{k}^\mathrm{d}=T_{k}^{\rm o}+T_{k}^{\rm c}$. Let $T_{k}^{\rm a}$, $\forall k\in\{0, \dots, K-2 \}$ represent the random number of time steps between two consecutive arrivals of the EB from operating periods $k$ and $k+1$ at $t_{k}$ and $t_{k+1}$, respectively. This corresponds to the sum duration of a pair of consecutive charging and operating periods, i.e., $T_{k}^{\rm a}=t_{k+1}-t_{k}=T_{k}^{\rm c}+T_{k+1}^{\rm o}$.  \par

We define $B_t$ as a random variable that represents the EB status at time step $t$, where $B_t=1$ and $B_t=0$ correspond to the charging and operating periods, respectively. In addition, let $k_t$ denote the index of the current charging or operating period at time step $t$ \footnote{To avoid potential confusion, we would like to highlight that $t_{k}$ is the index of a time step and $k_t$ is the index of a charging/operating period.}. \par

Let $\tau_t$ denote the number of remaining time steps from time step $t$ to the next departure time. The iterative calculation for $\tau_{t+1}$ is derived as 
\begin{equation}
\label{taut}
\\ \tau _{t+1}=\begin{cases}  T_{k_{t+1}}^\mathrm{d},&		\mathrm{if}\,\,B_{t+1}=0 \,\,\mathrm{and} \,\, B_{t}=1	\\ \tau _{t}-1,&		\mathrm{otherwise} \\\end{cases},
\end{equation}
\noindent which means that $\tau_t$ decreases by $1$ at each time step until it reaches $0$. When $\tau_t=0$, the EB will depart for operating period $k_{t+1}$ at the next time step $t+1$. In this case, $\tau_{t+1}$ is set to a fixed value $T_{k_{t+1}}^\mathrm{d}$ according to the bus schedule. 

We use $\varGamma_{t}(B_t)$ to denote the conditional probability that the current period is terminated at time step $t$, given whether the current period is charging (i.e., $B_t=1$) or operating (i.e., $B_t=0$). Note that $\varGamma_{t}(B_t)$ depends on $\tau_{t}$ and $k_t$. Intuitively, in the operating period ($B_t=0$), the probability of the EB returning to the terminal station increases as its traveling time elapses. Mathematically, we have
\begin{align}
\label{Gammat1}
\varGamma_{t}(B_t=0)&=\mathrm{Pr}(B_{t+1}  = 1|B_{t} = 0,\tau_{t},k_t)  \IEEEnonumber\\
&=\frac{\mathrm{Pr}(T_{k_t}^{\rm o}=T_{k_t}^\mathrm{d}-\tau_{t})}{\prod_{x=0}^{T_{k_t}^\mathrm{d}-\tau_{t}-1}[1-\mathrm{Pr}(T_{k_t}^{\rm o}=x)]}, 
\end{align} 
\noindent where the numerator represents the probability of the EB arriving at the terminal station at time step $t$, while the denominator represents the probability of the EB not arriving at the terminal station before time step $t$. \par

For the charging period ($B_t=1$), we consider that the bus line works in accordance with a fixed schedule and the EBs must depart on time. Therefore, when the scheduled departure time is reached, i.e., $\tau_t=0$, the charging period terminates with $\varGamma_{t}=1$. Meanwhile, we have $\varGamma_{t}=0$ in all the other time steps. Therefore, we have

\begin{align}
\label{Gammat2}
\varGamma_{t}(B_t=1)=\mathrm{Pr}(B_{t+1}  = 0|B_{t} = 1,\tau_{t}) 
=\begin{cases}  1,&		\mathrm{if}\,\,\tau_{t}=0 \\ 0,&		\mathrm{otherwise} \\\end{cases}. 
\end{align}

The optimization objective is to determine the EB charging schedule in the charging period that minimizes the charging cost while keeping a sufficient SoC level in the battery to ensure reliable operation in the operating period. The charging cost depends on real-time electricity prices, which fluctuate over time. At each time step $t$, we use $P_t$ to denote the electricity price.

Let $E_t$ denote the SoC level of the battery for the EB at time step $t$, which is constrained by 
\begin{equation}
\label{eq3}
E_{\min}\leqslant E_t\leqslant E_{\max}.
\end{equation}
Furthermore, let $C_t$ denote the charging/discharging power of the EB battery at time step $t$, which is positive when charging and negative when discharging. Note that the charging schedule $C_t$ only has to be determined in the charging period, where the EB can either draw energy from or return energy back to the electric grid in the V2G mode. On the other hand, the EB is always discharging in the operating period when the EB travels along the route, and thus $C_t$ is negative. Expressed mathematically, the constraint of $C_t$ can be written as: 
\begin{equation}
\label{ct_constraint}
C_t \in \begin{cases}
\left[ -D_{\max},C_{\max} \right]\cap \left[ \frac{E_{\min}-E_t}{\varDelta t},\frac{E_{\max}-E_t}{\varDelta t} \right] ,&		\mathrm{if}\,\,B_t=1\\
\left[ -D_{\max}, 0 \right]\cap \left[ \frac{E_{\min}-E_t}{\varDelta t},0 \right] ,&		\mathrm{if}\,\,B_t=0\\
\end{cases},
\end{equation}
\noindent where $C_{\max}$ and $D_{\max}$ denote the maximum absolute value of charging and discharging power, respectively. In addition, $\left[ \left( E_{\min}-E_t \right) /\varDelta t,\left( E_{\max}-E_t \right) /\varDelta t \right]$ represents the value range due to the limitation of the EB battery capacity. Finally, the dynamics of the EB battery can be modeled as
\begin{equation}
\label{Etdynamic}
E_{t+1}=E_{t}+ C_t \Delta t .
\end{equation}

\section{MDP Model}

\subsection{State}
Let $S_{t}=\left(E_{t}, B_t, \tau_t, H_t, k_t\right)$ be the system state at time step $t$. $H_t$ denotes the historical electricity prices in the period spanning from time step $(t-w_p)$ up to time step $t$, i.e.,
\begin{equation}
\label{price}
H_t=\left( P_{t-w_p}, P_{t-w_p+1},...,P_{t-1}, P_t \right) .
\end{equation}   
Note that $w_p$ is the length of the time window used for considering past prices. \par 
Let $\mathcal{S}^{+}$ denote the state space, which can be divided into the set of non-terminal states $\mathcal{S}=\{S_{t}|E_{t}\geq E_{\min}\}$ and the set of terminal states $\mathcal{S}^{\mathrm{T}}=\{S_{t}|E_{t}< E_{\min}\}$. When the SoC level in the EB's battery is lower than the minimum battery capacity constraint, i.e., $E_{t}< E_{\min}$, the agent will enter the terminal states and the current episode will end before the maximum time step $T$ is reached. 

\subsection{Action}
Let $A_t=C_t\in\mathcal{A}$ be the action at time step $t$, where $\mathcal{A}$ represents the action space. The charging scheduling policy determines the action $A_t$ only in those states $S_t$ that are associated with $B_t=1$. The action space $\mathcal{A}$ can be derived based on the first case in \eqref{ct_constraint}. When the EB is in the operating period ($B_t=0$), $C_t$ is a random variable whose value at each time step $t$ is given by the environment rather than being determined by the agent. The range of the random variable $C_t$ is specified by the second case in \eqref{ct_constraint}. \par

\subsection{Transition Probability}
The state transition probability is derived as
\begin{equation}
\begin{split}
\mathrm{Pr}\left( S_{t+1}|S_t,A_t \right) =\mathrm{Pr}\left( B_{t+1}|B_t,\tau_t,k_t \right)\mathrm{Pr}\left( k_{t+1}|k_t,B_t,B_{t+1} \right)\\
\mathrm{Pr}\left( \tau_{t+1}|\tau_t, B_t, B_{t+1} ,k_{t+1}\right)\mathrm{Pr}\left(E_{t+1}|E_t,A_{t} \right)\mathrm{Pr}\left( H_{t+1}|H_t \right),
\end{split}
\label{pr}
\end{equation}
\noindent where the transition probability of state $S_t$ is decomposed into the product of the transition probabilities of each component in the state definition. The transition probabilities of historical electricity prices $\mathrm{Pr}\left( H_{t+1}|H_{t} \right)$ are not available, but samples of the trajectory can be obtained from real-world data. The transition probability of SoC level, i.e., $\mathrm{Pr}\left( E_{t+1}|E_t,A_{t} \right)$ can be calculated by \eqref{Etdynamic}. Next, the transition probability $\mathrm{Pr}\left( B_{t+1}|B_t, \tau_t,k_t\right)$ can be derived from the termination probability $\varGamma_t(B_t)$, i.e.,
\begin{equation}
\label{prtaut}
\mathrm{Pr}\left( B_{t+1}|B_t, \tau_t,k_t\right)=\begin{cases}	1-\varGamma_t(B_t) ,&		\mathrm{if}\,\,B_{t+1}= B_{t}\\	\varGamma_t(B_t),&		\mathrm{if}\,\,B_{t+1}= 1-B_{t} \\\end{cases},
\end{equation}
\noindent where $\varGamma_t(B_t)$ is given by \eqref{Gammat1} and \eqref{Gammat2}. The transition probability of $k_t$, i.e., $\mathrm{Pr}\left( k_{t+1}|k_t,B_t,B_{t+1} \right)$, is derived by the iterative calculation for $k_{t+1}$, i.e.,
\begin{equation}
\label{dynamics of k_t}
\\ k _{t+1}=\begin{cases}  k_t+1,&		\mathrm{if}\,\,B_{t+1}=0 \,\,\mathrm{and} \,\, B_{t}=1	\\ k_t,&		\mathrm{otherwise} \\\end{cases}, 
\end{equation}
\noindent where $k_0=0$. Finally, the iterative calculation for $\tau_{t+1}$ in \eqref{taut} can be used for calculating $\mathrm{Pr}\left( \tau_{t+1}|\tau_t, B_t, B_{t+1},k_{t+1} \right)$. 

\subsection{Reward Function}
The optimization objective is to minimize the charging costs, while guaranteeing sufficient energy in the EB battery during the operating period. Therefore, we define the reward function as 
\begin{align}
\label{c}
r(S_t,A_t)=\begin{cases}	-C_{t}\Delta t P_t B_t ,&		\mathrm{if}\,\,S_{t}\in \mathcal{S}\\	
-C^{\mathrm{end}},&		\mathrm{if}\,\,S_{t}\in \mathcal{S}^{\mathrm{T}} \\\end{cases},
\end{align}
\noindent where $C_{t}\Delta t P_t B_t$ is the charging cost and $C^{\mathrm{end}}$ is the penalty to the agent for entering the terminal states. This means that the EB battery is not sufficiently charged to support the entire operating period to completion. As a result, the EBs having depleted batteries receive more negative expected cumulative rewards than their counterpart associated with normal operation, due to receiving a large negative reward $-C^{\mathrm{end}}$ upon entering the terminal state. In practice, $C^{{ \rm end}}$ is a hyperparameter tuned empirically. If set too small, the agent risks reaching the terminal state, whereas a value that is too large may lead to an overly conservative policy, increasing charging costs. In our experiments, the optimal $C^{{ \rm end}}$ is typically around $5$ to $10$ times $C_t \Delta t P_t B_t$, ensuring that the agent consistently avoids terminal states while still optimizing for lower charging cost.\par

\subsection{Optimization Objective}

Define the return $G_t$ as the sum of rewards starting from time step $t$, i.e.,
\begin{equation}
\label{cumulative reward}
G_t=\sum_{t'=t}^{T-1}{r\left( S_{t'},A_{t'} \right)}.
\end{equation}

The state-value or value function $V_{\pi}(s)$ is defined as the expected return from starting in state $s$ and then following policy $\pi$, i.e.,
\begin{equation}
\label{value}
V_{\pi}(s)=\mathbb{E}_{\pi}\left[G_t\arrowvert S_t=s\right], \forall s\in\mathcal{S}.
\end{equation}

\noindent Note that the value function of terminal states is always $-C^{\rm end}$, i.e., $V_{\pi}(s)=-C^{\rm end},\forall s\in\mathcal{S}^{\mathrm{T}}$. Thus, our objective is to derive the optimal policy $\pi^*$ that maximizes the value function for all the non-terminal states
\begin{equation}
\label{ori 3}
V_{\pi^{*}}(s)=\max_{\pi}V_{\pi}(s), \forall s\in\mathcal{S}.
\end{equation}

In order to derive the optimal policy, the action-value function or Q function $Q_{\pi}(s,a)$ is defined as the expected return from starting in state $s$, taking action $a$, and then following policy $\pi$, i.e.,
\begin{equation}
\label{q}
Q_{\pi}(s,a)=\mathrm{E}_{\pi}\left[G_t\arrowvert S_{t}=s,A_{t}=a\right], \forall s\in\mathcal{S}, a\in\mathcal{A}.
\end{equation}

\noindent Note that the Q function for the terminal states is always $-C^{\rm end}$, i.e., $Q_\pi(s,a)=-C^{\rm end}, \forall s\in\mathcal{S}^{\mathrm{T}},a\in\mathcal{A}$. The optimal action-value function is then $Q_{\pi}^{*}(s,a)=\max_{\pi}Q_{\pi}(s,a)$. Given $Q_{\pi}^{*}(s,a)$, the optimal policy selects the highest-valued action in each state.\par 

In order to estimate $Q_{\pi}^{*}(s,a)$ typically, the following Bellman Equation is used as an iterative update, where
\begin{equation}
\label{bellman}
\begin{split}
Q^{*}\left( s,a \right) = \mathbb{E}\left[ r\left( s,a \right) +\max_{a'}Q^{*}(s',a')  \right] .
\end{split}
\end{equation}

The formulated EB charging problem corresponds to an episodic task with a maximum of $T$ time steps in an episode. An episode terminates earlier if the terminal state is reached, which means that the EB runs out of battery. For any full episode where the EB does not run into the terminal state, the last period is always the charging period $K-1$, where the EB has to charge sufficient energy for the operating period $0$ of the next day. The optimal policy is stationary for any episodic task \cite{sutton1999between,puterman2014markov}.

\section{Options over MDP}

The MDP model defined in Section IV is essentially a long-range multiple-phase planning problem, since the time horizon is divided into multiple charging and operating periods, where each phase consists of a large number of time steps. In the following, we adopt the framework of options \cite{sutton1999between,bacon2017option} to tackle this problem, which enables the agent to abstract actions at different temporal levels. \par       

\subsection{Definition of Options}

Let $\omega \in \varOmega $ denote the options, where $\varOmega$ is the option space. Options can be regarded as temporally extended ``actions", which can last for multiple time steps. An option is prescribed by the \emph{policy over options} $\mu$ according to the current state $S_{t}$. Each option $\omega$ is associated with the triple of $\left( \mathcal{I}_\omega , \pi_\omega , \beta_\omega \right)$, where $\mathcal{I}_\omega $ is the initiation set of states, i.e., $\omega$ is available in state $S_t$ if and only if $S_t \in \mathcal{I}_\omega$, while $\pi_\omega$ is the \emph{intra-option policy} and $\beta_\omega$ is the termination condition. If an option $\omega$ is chosen at time step $t$, the option will terminate with probability $\beta_\omega(S_{t+l})$ at each time step $t+l$, $\forall l\in{1,2,\dots}$. If the option $\omega$ does not terminate at time step $t+l$, $\forall l\in{0,1,2,\dots}$, an action $A_{t+l}$ is chosen according to the intra-option policy $\pi_\omega(S_{t+l})$ and the environment moves on to the next time step $t+l+1$. Otherwise, a new option $\omega'$ may be selected according to the policy over options $\mu(S_{t+l})$, and the same process is repeated for the following time steps.\par

In this paper, we define the charging target option $\omega$, which corresponds to the target SoC level of the EB's battery, when a charging period terminates. Note that the option space is defined by the $E_t$ constraints in \eqref{eq3}. Since the charging targets only have to be determined in the charging periods, $\mathcal{I}_\omega$ is defined as
\begin{equation}
\label{initI}
\mathcal{I}_\omega=\left\lbrace S_t\in \mathcal{S}|B_t=1\right\rbrace .
\end{equation}
\noindent The termination condition is defined as
\begin{equation}
\label{termI}
\beta_\omega(S_t)=\left\{
\begin{array}{ll}
1, & \mathrm{if} \ B_{t}=0\,\,\mathrm{and}\,\, B_{t+1}=1\\
0, & \mathrm{otherwise}  \\
\end{array}\right. ,
\end{equation} which means the option should be terminated when an operating period ends and the next charging period will start. 
\subsection{Decoupling the original MDP}
\begin{figure}[t]
\centering
\includegraphics[width=0.57\linewidth]{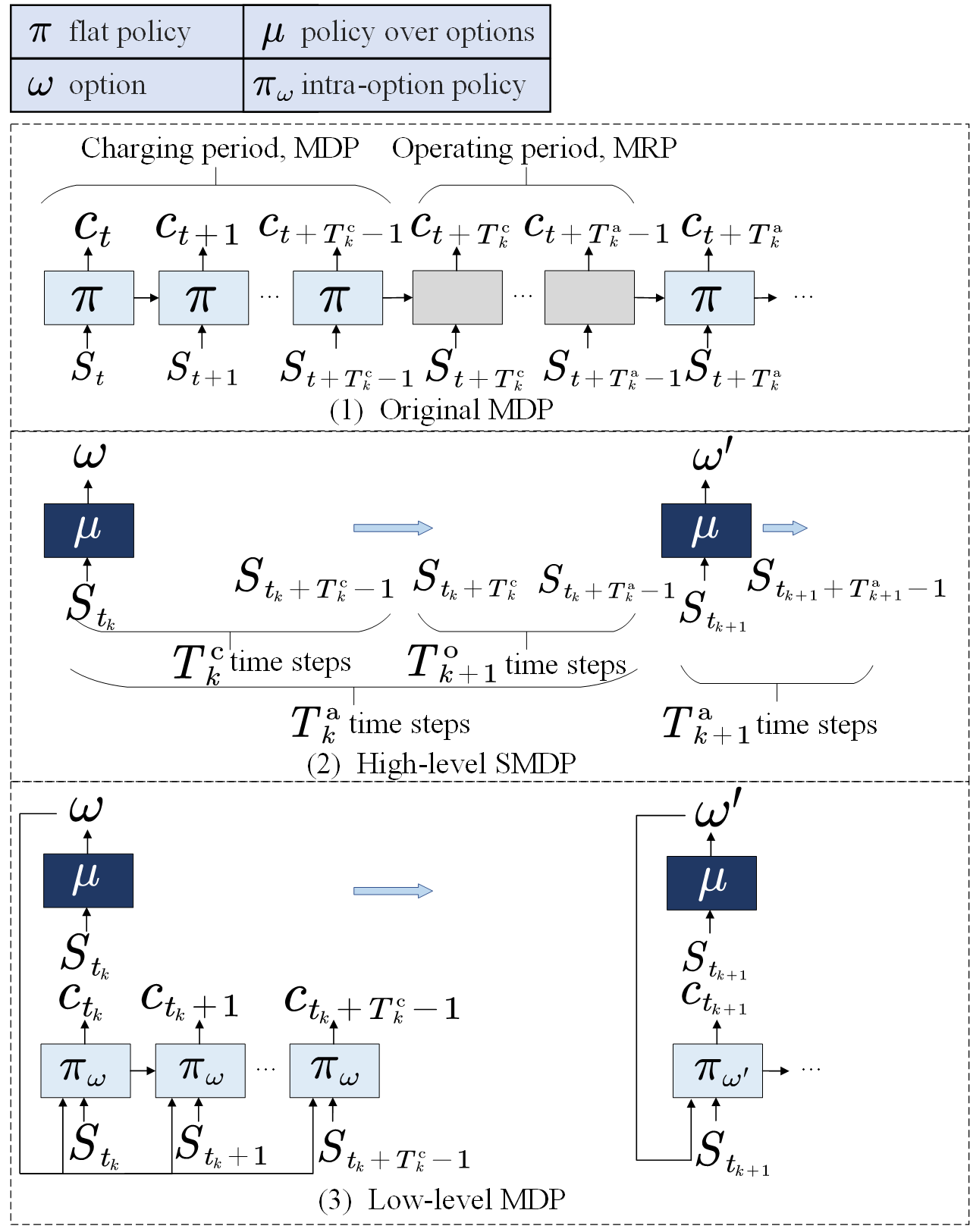}
\caption{The schematic diagram of the original MDP, high-level SMDP, and low-level MDPs.}
\label{two-level}
\end{figure}

As shown in Fig.\ref{two-level}, by introducing options, our original MDP model is divided into low-level and high-level decision processes. The top panel illustrates the original MDP defined in Section IV, where a flat policy $\pi$ is used for determining the charging schedule action $C_{t}$ at each time step $t$ during the charging periods. Meanwhile, $C_{t}$ is given by the environment according to the amount of power discharged during each time step in the operating periods. Therefore, the original MDP consists of an interleaving of MDPs corresponding to charging periods and Markov Reward Processes (MRPs) representing the operating periods. 

Although the original MDP can be solved directly to obtain $\pi$, a more efficient solution is to divide it into two levels through options. The middle panel shows the high-level SMDP, where the policy over options $\mu$ prescribes the charging target option $\omega$ at the selected time steps $\{t_{k}\}_{k=0}^{K-2}$ when the charging periods start. Assume that option $\omega$ is selected according to $\mu(S_{t_{k}})$ at time step $t_{k}$ when a charging period starts. When the charging period terminates after a random number of $T_{k}^{\rm c}$ time steps, the state $S_{t_{k}}$ transits to $S_{t_{k}+T_{k}^{\rm c}}$, where $E_{t_{k}+T_{k}^{\rm c}}$ should be as close to the desired charging target $\omega$ as possible. When the next charging period starts at time step $t_{k+1}=t_k+T_{k}^{\rm a}$, the next option $\omega'$ is selected according to $\mu(S_{t_{k+1}})$. 

Finally, the bottom panel shows that the low-level MDPs are embedded into the high-level SMDP, where at each charging period, the intra-option policy $\pi_\omega$ of the selected option $\omega$ determines the optimal charging schedule action $C_{t}$ that minimizes the charging cost, while ensuring that the charging target of $\omega$ is realized. By superimposing the high-level policy over options $\mu$ and the low-level intra-option policy $\pi_\omega$, a flat policy is constructed to solve the original MDP problem. Notably, the long time-horizon of the original MDP is divided into multiple shorter time-horizons of the low-level MDPs. Since each low-level MDP is provided with a charging target option obtained by solving the high-level MDP, the low-level MDPs become independent of each other. In other words, each low-level MDP only has to concentrate on its immediate time-horizon for decision-making, without having to look further into the future and consider those low-level MDPs in subsequent time-horizons. 

By adopting the hierarchical structure and solving the high-level SMDP and low-level MDPs instead of the original MDP, the learning efficiency and convergence speed can be significantly improved. \par  

\subsection{The High-Level SMDP}
When we only consider the high-level policy over options $\mu$ to select options (i.e., assuming the intra-option policy $\pi_{\omega}$ is available for each option $\omega\in\Omega$), the decision process becomes an SMDP \cite{puterman2014markov} that can be derived from the original MDP.\par 

In SMDP, state transitions and action selections take place at discrete time, but the length of time interval from one transition to the next is random. Therefore, an SMDP consists of state space, action space, an expected cumulative discounted reward for each state-action pair, and a joint distribution of the next state and transit interval. In our high-level SMDP, the states are the same as those of the original MDP, while the actions correspond to the options. For each state transition when option $\omega$ is selected, the transit time is ${T}_k^{\omega}$ time steps. Note that an option terminates when either the EB arrives at the terminal station where $T_{k}^{\omega}=T_{k}^{\rm a}$, or the EB runs out of battery during the trip where $T_{k}^{\omega}< T_{k}^{\rm a}$. The SMDP model has its corresponding reward $r^{\mathrm{H}}\left( s,\omega \right)$, i.e., the expected cumulative reward within one transition of SMDP in ${T}_k^{\omega}$ time steps. Let $\mathcal{E}(\omega,s,t_k)$ denote the event that $\omega$ starts in state $s$ at time step $t_k$. The reward $r^{\mathrm{H}}\left( s,\omega \right)$ can be derived by
\begin{equation}
\label{optionreward}
r^{\mathrm{H}}\left( s,\omega \right) =\mathbb{E}_{\pi_{\omega}} \left[\sum_{l=0}^{{T}_k^{\omega}-1}{r(S_{t_k+l},A_{t_k+l}) }|\mathcal{E}(\omega,s,t_k)\right],
\end{equation}
\noindent where the action $A_{t_k+l}$ is derived based on the intra-option policy $\pi_{w}$ for the first $T_{k}^{\rm c}$ time steps during the charging period, and then provided by the environment for the following $T_{k}^{\omega}-T_{k}^{\rm c}$ time steps during the operating period.  \par 

The transition distributions of the SMDP $P_{ss'}^{\omega}$ can be formulated as 
\begin{equation}
\label{Pss'}
P_{ss'}^{\omega}=\sum_{{T}_k^{\omega}=1}^{\infty}{P\left(s',{T}_k^{\omega} \right)},
\end{equation}
\noindent where $P(s',{T}_k^{\omega})$ represents the probability that the option terminates after ${T}_k^{\omega}$ time steps when the state $s$ changes to $s'$.

Let us define the high-level return $G_{k}^{\rm H}$ as the sum of the high-level rewards starting from time step $t_k$, i.e.,
\begin{align}
\label{Gt^H}
G_{k}^{\rm H}=\sum_{k'=k}^{K-1}{r^{\rm H}\left( S_{t_{k'}},\omega_{t_{k'}} \right)}, \forall k\in\{0,1,\cdots,K-2\}
\end{align}

The high-level value function $V^{\rm H}_{\mu}(s)$ is defined as the expected high-level return from starting in state $s$ and then following the policy over options $\mu$, i.e.,
\begin{equation}
\label{value_high}
V^{\rm H}_{\mu}(s)=\mathbb{E}_{\mu}\mathbb{E}_{\pi_{\omega}}\left[	G_{k}^{\rm H}\arrowvert S_{t_k}=s\right], \forall s\in\mathcal{S}.
\end{equation}

\noindent Note that the high-level value function of the terminal states is always $-C^{\rm end}$, i.e., $V^{\rm H}_{\mu}(s)=-C^{\rm end},\forall s\in\mathcal{S}^{\mathrm{T}}$. The objective of the high-level SMDP is to derive the optimal policy over option $\mu^*$ that maximizes the high-level value function for all the nonterminal states
\begin{equation}
\label{value_high 2}
V^{\rm H}_{*}(s)=V^{\rm H}_{\mu^{*}}(s)=\max_{\mu}V^{\rm H}_{\mu}(s), \forall s\in\mathcal{S}.
\end{equation}

In order to derive the optimal policy over options, the option-value function or high-level Q function $Q_{\mu}^{\rm H}(s,\omega)$ is defined as the expected return from starting in state $s$, taking option $\omega$, and then following policy $\mu$, i.e.,
\begin{equation}
\label{qH}
Q_{\mu}^{\rm H}(s,\omega)=\mathrm{E}_{\mu}\mathbb{E}_{\pi_{\omega}}\left[	G_t^{\rm H}  \arrowvert S_{t_k}=s,\omega_{t_k}=\omega\right],   \forall s\in\mathcal{S}, \omega\in\mathcal{\varOmega}.
\end{equation}

\noindent Note that the option-value function for the terminal states is always $-C^{\rm end}$, i.e., $Q_\mu^{\rm H}(s,\omega)=-C^{\rm end}, \forall s\in\mathcal{S}^{\mathrm{T}},\omega\in\mathcal{\varOmega}$. Given the optimal option-value function $Q^{\rm H}_{*}(s,\omega)=\max_{\mu}Q_{\mu}^{\rm H}(s,\omega)$, the optimal policy over options selects the highest-valued option in state $s$ at time step $t_k$. The Bellman Optimality Equation is 
\begin{equation}
\label{bellmanQH}
\begin{split}
Q_{*}^{\rm H}\left( s,\omega \right) = \mathbb{E}\left[ r^{\rm H}\left( s,\omega \right) +\max_{\omega'}Q^{\rm H}_{*}(s',\omega')  |\mathcal{E} \left( \omega,s \right) \right] ,
\end{split}
\end{equation}

\noindent where $\mathcal{E} \left( \omega,s \right)$ denotes option $\omega$ being initiated in state $s$. \par

\subsection{The Low-Level MDP}
As illustrated in Fig. \ref{two-level}, the course of the original MDP is divided into an interleaving of low-level MDPs and MRPs, where the time horizon $T_{k}^{\rm c}$ (resp. $T_{k}^{\rm o}$) of each low-level MDP (resp. MRP) corresponds to the charging (resp. operating) period in one transition of the high-level SMDP. In order to derive the intra-option policy $\pi_{\omega}$, the low-level MDP is formulated below. \par
In the low-level MDPs, the state is denoted as $(S_t,\omega_t)$, where the state in the original MDP, i.e., $S_t$, is augmented by including the current option $\omega_t$. Let $\mathcal{S}_{\rm L}^{+}$ denote the low-level state space, which can be divided into the set of low-level non-terminal states $\mathcal{S}_{\rm L}=\{S_{t}\in\mathcal{S},\omega_t\in\varOmega|B_{t}=1\}$ and the set of low-level terminal states $\mathcal{S}_{\rm L}^{\mathrm{T}}=\{S_{t}\in\mathcal{S},\omega_t\in\varOmega|B_{t}=0\}$. Note that the low-level MDP terminates whenever the EB enters the operating period, i.e., $B_{t}=0$. The actions and the transition probabilities are the same as those of the original MDP, since $\omega_t$ remains the same throughout the time horizon of the low-level MDP. Since we should ensure that the charging target of the selected option $\omega$ is realized at the end of the current charging period, the intrinsic reward of the low-level MDPs is defined as

\begin{equation}
\label{rewardL}
r^{\mathrm{L}}(s,\omega,a) =\begin{cases}
-{C_t}\Delta tP_tB_t,&		\mathrm{if}\,\,(s,\omega)\in \mathcal{S}_{\rm L}\\
- \kappa {c^{\mathrm{tar}}(s,\omega)}  ,&		\mathrm{if}\,\,(s,\omega)\in \mathcal{S}_{\rm L}^{\mathrm{T}}\\
\end{cases},
\end{equation}

\noindent where
\begin{equation}
\label{target_mse}
c^{\mathrm{tar}}(s,\omega)=\left( \omega -E_{t_k+T_k^\mathrm{c}} \right) ^2.
\end{equation}
\noindent Note that $s=S_t$ and $a=\pi_\omega(s)$ when $(s,\omega)\in \mathcal{S}_{\rm L}$. Although no action is selected when $(s,\omega)\in \mathcal{S}_{\rm L}^{\rm T}$, we keep the action in the reward function of the terminal state for notational simplicity. Furthermore, $c^{\mathrm{tar}}(s,\omega)$ in \eqref{target_mse} represents the penalty of not realizing the charging target, which is the squared error between the charging target $\omega$ and the actual SoC level $E_{t_k+T_k^\mathrm{c}}$ at the end of the charging period. Finally, $\kappa$ in \eqref{rewardL} is the coefficient representing the relative importance of minimizing the charging cost versus the penalty $c^{\mathrm{tar}}(s,\omega)$. $\kappa$ is set to a high enough value to ensure that the charging target is met whenever possible. \par

Let us define the low-level return $G^{\rm L}_t$ as the sum of the intrinsic rewards from time step $t$ to the end of the time horizon of the low-level MDP, i.e.,
\begin{align}
\label{Gt^L}
G_t^{\rm L}=\left[\sum_{t'=t}^{{t_{k}+T_k^{\rm c}}}{r^{\rm L}\left( S_{t'},\omega,A_{t'} \right)} \right], \forall t\in[t_{k},t_{k}+T_k^{\rm c}]. 
\end{align}

The low-level value function $V^{\rm L}_{\pi_\omega}(s,\omega)$ is defined as the expected return from starting in state $(s,\omega)$ and then following the intra-option policy $\pi_\omega$, i.e.,
\begin{equation}
\label{value_low}
V^{\rm L}_{\pi_\omega}(s,\omega)=\mathbb{E}_{\pi_\omega}\left[G_t^{\rm L}
\arrowvert S_t=s,\omega \right], \forall (s,\omega)\in\mathcal{S}_{\rm L}. 
\end{equation}

\noindent Note that the value function of low-level terminal states is always $- \kappa c^{\mathrm{tar}}(s,\omega)$, i.e., $V^{\rm L}_{\pi_\omega}(s,\omega)=- \kappa c^{\mathrm{tar}}(s,\omega),\forall (s,\omega)\in\mathcal{S}_{\rm L}^{\mathrm{T}}$. The objective of the low-level MDP is to derive the optimal intra-option policy $\pi_\omega^*$ that maximizes the low-level value function for all the states
\begin{equation}
\label{optimal low value}
V^{\rm L}_{*}(s,\omega)=V^{\rm L}_{\pi_\omega^*}(s,\omega)=\max_{\pi_\omega}V^{\rm L}_{\pi_\omega}(s,\omega), \forall (s,\omega)\in\mathcal{S}_{\rm L}.
\end{equation}

In order to derive the optimal intra-option policy, the low-level action-value function or Q function $Q_{\pi_\omega}^{\rm L}(s,\omega,a)$ is defined as the expected low-level return from starting in state $(s,\omega)$, taking action $a$, and then following policy $\pi_\omega$, i.e.,
\begin{equation}
\label{qL}
\begin{split}
Q_{\pi_\omega}^{\rm L}(s,\omega,a)=\mathrm{E}_{\pi_\omega}\left[G_t^{\rm L}\arrowvert S_{t}=s,A_{t}=a,\omega \right],\\
\forall (s,\omega)\in\mathcal{S}_{\rm L}, a\in\mathcal{\mathcal{A}}.
\end{split}
\end{equation}

Note that the action-value function for the low-level terminal states is always $- \kappa {c^{\mathrm{tar}}(s,\omega)}$, i.e., $Q_{\pi_\omega}^{\rm L}(s,\omega,a)=- \kappa {c^{\mathrm{tar}}(s,\omega)}, \forall (s,\omega)\in\mathcal{S}_{\rm L}^{\mathrm{T}},a\in\mathcal{\mathcal{A}}$. Given the optimal action-value function $Q^{\rm L}_{*}(s,\omega,a)=\max_{\pi_\omega}Q_{\pi_\omega}^{\rm L}(s,\omega,a)$, the optimal intra-option policy selects the highest-valued action in state $(s,\omega)$. The  Bellman Optimality Equation is given as
\begin{equation}
\label{bellmanQL}
\begin{split}
Q_{*}^{\rm L}\left( s,\omega,a \right) = \mathbb{E}\left[ r^{\rm L}\left( s,\omega,a \right) +\max_{a'}Q^{\rm L}_{*}(s',\omega,a')  \right] .
\end{split}
\end{equation}



\newtheorem{thm}{Theorem}
\begin{thm}
\label{thm_1}
The flat policy created by superimposing the optimal high-level policy over options $\mu^*$ and the optimal low-level intra-option policy $\pi_\omega^*$ performs as well as the optimal policy of the original MDP $\pi^{*}$, i.e.,

\begin{equation}
\label{thm}
V_{\pi^*}(s)=\mathbb{E} _{{\mu^*}}\mathbb{E} _{\pi_{\omega^*}^*}\left[\sum_{t=0}^{T-1} {r\left( S_{t},A_{t} \right)} |S_{0}=s \right], \forall s\in\mathcal{S},
\end{equation} 	

\noindent where $\omega^*=\mu^*(s)$, $\forall s\in\mathcal{S}$ represents the charging targets prescribed by $\mu^*$. Moreover, $\mu^*$ is the optimal high-level policy over options when the optimal low-level intra-option policy $\pi_\omega^*$ is available in the high-level SMDP.
\end{thm}

\section{DRL Solution}
In this section, we present a DRL algorithm, namely the HDDQN-HER conceived for learning the optimal policy over options $\mu^*$ by solving the high-level SMDP; and the optimal intra-option policy $\pi_\omega^*$ by solving the low-level MDPs. 

\subsection{Overall Training Framework}
\begin{figure}[t]
\centering
\includegraphics[width=0.5\linewidth]{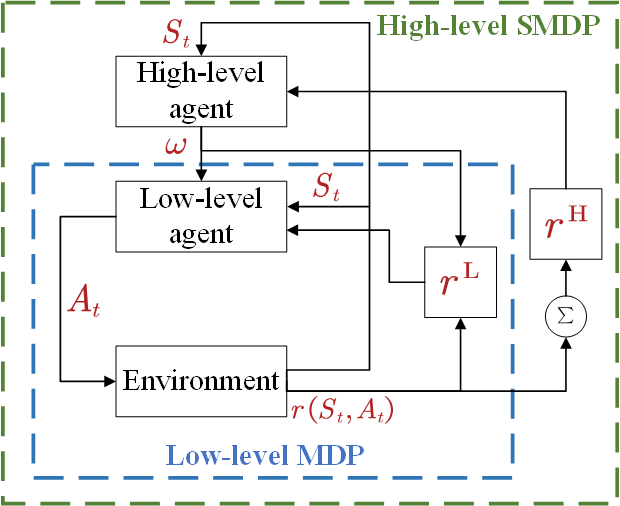}
\caption{The framework of HDDQN-HER algorithm.}
\label{flow}
\end{figure}

We adopt two deep Q networks, i.e., $\mathcal{Q}^\mathrm{H}\left( \cdot ;\theta^\mathrm{H} \right)$ and $\mathcal{Q}^\mathrm{L}\left( \cdot ;\theta^\mathrm{L} \right)$ with parameters $\theta^\mathrm{H}$ and $\theta_\mathrm{L}$ to approximate $Q^{\mathrm{H}}_*\left( s,\omega \right)$ and $Q^{\mathrm{L}}_*\left( s,\omega,a \right)$, respectively. The two networks are jointly trained, where the overall framework is illustrated in Fig. \ref{flow}. At each time step $t$, both the high-level and low-level agents receive state $S_{t}$ from the environment. The high-level agent only selects a charging target option $\omega_t$ based on $S_{t}$ when a new charging period starts, i.e., $B_{t}=1, B_{t-1}=0$. The option is fed to the low-level agent instead of being fed to the environment. The low-level agent selects a charging schedule action $A_{t}$ at every time step $t$ in the charging period ($B_{t}=1$) based on the state $S_{t}$ and the option $\omega_t$. The action $A_{t}$ is fed to the environment, which responds to the agent by providing the reward $r(S_{t},A_{t})$. Meanwhile, the environment directly provides the agent with the reward at each time step $t$ in the operating period  ($B_{t}=0$) without requiring an action from the agent. The high-level agent calculates the high-level reward $r^\mathrm{H}(S_t,\omega_t)$ by summing up the reward $r(S_t,A_t)$ during each SMDP transition. Meanwhile, the low-level agent derives the low-level reward $r^\mathrm{L}(S_t,\omega_t,A_t)$ at each time step based on $r(S_t,A_t)$ and $\omega_t$ according to \eqref{rewardL} and \eqref{target_mse}. \par   

The two Q networks $\mathcal{Q}^\mathrm{H}\left( \cdot ;\theta^\mathrm{H} \right)$ and $\mathcal{Q}^\mathrm{L}\left( \cdot ;\theta^\mathrm{L} \right)$ are trained simultaneously based on the DDQN algorithm. We refer interested readers to \cite{van2016deep} for details of the DDQN algorithm. The high-level and low-level agents draw experiences from their own replay buffers, i.e., $\mathcal{R}^\mathrm{H}$ and $\mathcal{R}^\mathrm{L}$, respectively. The two replay buffers are updated at different time scales. More specifically, transitions for the low-level MDPs are collected and stored into $\mathcal{R}^\mathrm{L}$ at each time step in the time horizon of the low-level MDP. However, only when the agent enters the terminal state or when an operating period terminates (i.e., $B_{t}=0$ and $B_{t+1}=1$), will a new transition for the high-level SMDP be collected and stored into $\mathcal{R}^\mathrm{H}$.  \par

\subsection{Overcoming Non-Stationarity in Training High-Level Agent}
Based on the above framework, it is challenging to directly learn the high-level policy over options $\mu^*$ and low-level intra-option policy $\pi_\omega^*$ in parallel. This is mainly due to the non-stationary environment of the high-level agent, which is caused by exploring and updating low-level policy during training. Specifically, the non-stationarity has two facets.

\begin{itemize}
\item Non-stationary reward function: according to the low-level reward function defined in \eqref{rewardL}, the low-level agent learns to minimize the cumulative charging cost during the charging period, while ensuring that the charging target is realized. Therefore, the cumulative charging cost decreases, while the low-level policy improves during training, which means that the high-level reward function $r^{\mathrm{H}}(s,\omega)$ as defined in \eqref{optionreward} changes over time.
\item Non-stationary transition probabilities: a sub-optimal low-level policy activated during training might not be able to achieve the charging target $\omega$ set by the high-level agent, while the charging target is actually achievable by the optimal low-level policy. In this case, the transition probability $P^{\omega}_{ss'}$ of the high-level SMDP will change over time as the low-level policy learns to achieve the charging target.  
\end{itemize}

To overcome the above challenges, we adopt the following pair of measures so that the high-level SMDP model remains independent of the changing and exploring low-level policy.

\begin{itemize}
\item To deal with the non-stationary reward function, we divide the joint learning process into two phases. In the first phase when the number of training episodes is smaller than a threshold, i.e., $e < M_\epsilon$, the high-level rewards $\hat{r}^{\mathrm{H}}$ stored in the high-level replay buffer $\mathcal{R}^\mathrm{H}$ are obtained by running a fixed low-level policy $\pi_{\rm q}$ instead of those actually achieved by the low-level behavior policy that is being trained. In the second phase when $e \geqslant M_\epsilon$, we calculate the high-level rewards $\hat{r}^{\mathrm{H}}$ based on the low-level behavior policy, i.e., $\hat{r}^{\mathrm{H}}={r}^{\mathrm{H}}$, since the low-level policy almost converges to the optimal one at this phase and the non-stationary reward issue is negligible.  
\item To deal with the non-stationary transition probabilities, the actual SoC level of the battery $E_{t_k+T_{k}^{\rm c}}$ at the end of the charging period is stored in the high-level replay buffer $\mathcal{R}^\mathrm{H}$ as the option, instead of the actual option $\omega$ that has been prescribed by the high-level agent at time step $t_k$, when the charging period started. Therefore, the transition stored in $\mathcal{R}^\mathrm{H}$ is $\left[S_{t_k}, \tilde{\omega}, \hat{r}^{\mathrm{H}}(S_{t_k}, \tilde{\omega}), S_{t_k+{T}_k^{{\rm a}}}\right]$ instead of $\left[S_{t_k}, \omega, \hat{r}^{\mathrm{H}}(S_{t_k}, \omega), S_{t_k+{T}_k^{\rm a}}\right]$, where $\tilde{\omega}= E_{t_k+T_{k}^{\rm c}}.$ Note that the next state $S_{t_k+{T}_k^{{\rm a}}}$ is the one that is actually visited in this episode, and the reward $\hat{r}^{\mathrm{H}}(S_{t_k}, \tilde{\omega})$ is calculated as described above, depending on the current training phase. 
\end{itemize}


\subsection{Improving Sample Efficiency in Training the Low-Level Agent}
The low-level agent has to learn to achieve multiple charging targets, while only at the end of an episode will the penalty $c^{\mathrm{tar}}(s,\omega)$ of not realizing a given charging target be available in the low-level reward defined in \eqref{rewardL}. Both the multi-goal and sparse reward issues impair the sample efficiency in training the low-level agent, which leads to slow convergence in learning the low-level policy, as well as to an adverse effect in overcoming the non-stationarity for the high-level agent. To deal with the above challenges, we harness the following measures for improving the sample efficiency.

\begin{itemize}
\item \emph{Restricting the Option Space}: In each charging period, the low-level agent tries to achieve the charging target prescribed by the high-level agent. However, some charging targets might be unachievable, when the high-level agent has not yet learned a good policy. As a result, learning to achieve such charging targets is futile for the low-level agent, and the corresponding experiences are wasted.  
\begin{itemize}
\item Given the SoC level $E_t$ at the beginning of a charging period, some charging targets are unachievable by the end of the charging period even if the EB is fully charged or discharged at all the time steps. According to the dynamics of the SoC level in \eqref{Etdynamic}, a state-dependent option space $\varOmega'$ can be defined as
\begin{equation}
	\label{omegaspace'}
	\varOmega' = \left\lbrace \omega \in \varOmega | \omega \in \left[E_t-\tau_{t} D_{\max},E_t+\tau_{t} C_{\max}  \right] \right\rbrace .
\end{equation} 
\item If a charging target is too low, the EB will run out of battery power during operation and the MDP enters the terminal states. To exclude these unreasonable charging targets, all transitions $\left(\{-,\omega\}, -, -, \{-,\omega\} \right) $ associated with the current option $\omega$ are deleted from $\mathcal{R}^\mathrm{L}$, whenever the agent enters the terminal states. 
\end{itemize}
\item \emph{HER}: In addition to the original transition $\left(\{s,\omega\}, a, r^\mathrm{L}, \{s',\omega\} \right)$ that is stored in the low-level replay buffer $\mathcal{R}^\mathrm{L}$, we create another hindsight transition, i.e., $\left(\{s,TBD\}, a, \hat{r}^\mathrm{L}, \{s',TBD\} \right)$. At the end of the charging period, the actual achieved SoC levels $E_{t_k+T_{k}^{\rm c}}$ are filled in the TBD components and the additional hindsight transitions are stored into $\mathcal{R}^\mathrm{L}$ if the charging target $\omega$ is not realized. Note that $\hat{r}^\mathrm{L}={r}^\mathrm{L}$ at all the time steps, except for the last time step in the time horizon of the low-level MDP, i.e., $B_t=0$ and $B_{t-1}=1$, where $c^{\mathrm{tar}}(s,\omega)=0$ in $\hat{r}^\mathrm{L}(s,\omega,a)$ since the substituted charging target $E_{t_k+T_{k}^{\rm c}}$ is realized. This method guarantees that at least one sequence of transitions is stored in $\mathcal{R}^\mathrm{L}$ that succeeds in realizing the charging target for each charging period. 
\end{itemize}

Algorithm \ref{alg-H-DQN} describes the pseudocode of the HDDQN-HER algorithm. The functions to select actions or options, store hindsight transitions into $\mathcal{R}^\mathrm{L}$, and to update the associated Q networks are presented by Algorithms \ref{alg-HDQN-greedy}, \ref{alg-HDQN-store}, and \ref{alg-HDQN-update}, respectively. In Algorithm \ref{alg-HDQN-greedy}, the options or actions are prescribed based on the popular $\epsilon$-greedy policy, while in Algorithm \ref{alg-HDQN-update}, the two Q networks $\mathcal{Q}^\mathrm{H}\left( \cdot ;\theta^\mathrm{H} \right)$ and $\mathcal{Q}^\mathrm{L}\left( \cdot ;\theta^\mathrm{L} \right)$ are updated based on the corresponding sampled minibatch of transitions.\par

\begin{algorithm}[t]
\caption{HDDQN-HER algorithm}
\label{alg-H-DQN}
\begin{algorithmic}[1]  
\State Initialize $\mathcal{Q}^{\mathrm{H}}\left( s,\omega ;\theta^{\mathrm{H}} \right)$ and $\mathcal{Q}^{\mathrm{L}}\left( s,\omega,a ;\theta^{\mathrm{L}} \right)$ with $\theta^{\mathrm{H}}=\theta^{\mathrm{H}}_0$ and $\theta^{\mathrm{L}}=\theta^{\mathrm{L}}_0$, respectively. Initialize target Q networks $\mathcal{Q}^{\mathrm{H}-}\left( s,a ;\theta^{\mathrm{H}-} \right)$ and $\mathcal{Q}^{\mathrm{L}-}\left( s,\omega,a ;\theta^{\mathrm{L}-} \right)$ with $\theta^{\mathrm{H}-}\gets \theta^{\mathrm{H}} $ and $\theta^{\mathrm{L}-}\gets \theta^{\mathrm{L}} $. Initialize $\mathcal{R}^{\mathrm{H}}$ and $\mathcal{R}^{\mathrm{L}}$, exploration probability $\epsilon^{\mathrm{L}} = 1$ and $\epsilon^{\mathrm{H}} = 1$. 
\For{episode $e = 1, ..., M$}
\State Receive the start state $s$
\State Derive the option space $\varOmega'$ based on \eqref{omegaspace'}
\State $r^\mathrm{H}(s,\omega)\gets 0$, $s_0 \gets s$
\State $\omega\gets$ \Call{$\epsilon$-Greedy}{$s_0$,$\varOmega'$,$\epsilon^\mathrm{H}$,$\mathcal{Q}^{\mathrm{H}}\left( s_0,\omega ;\theta^{\mathrm{H}}\right)$}, $\tilde{\omega}\gets \omega$ 

\For{$t = 1, ..., T$}
\If{$B_t=1$}
\State $a\gets$ \Call{$\epsilon$-Greedy}{$\{s,\omega\}$,$\mathcal{A}$,$\epsilon^\mathrm{L}$,$\mathcal{Q}^{\mathrm{L}}\left( s,\omega,a ;\theta^{\mathrm{L}} \right)$} 
\State Execute action $a$ and observe reward $r(s,a)$, and next state $s'$ from environment
\State Derive low-level reward $r^\mathrm{L}(s,\omega,a)$,
\State Store $\left(\{s,\omega\}, a, r^\mathrm{L}, \{s',\omega\} \right) $ into $\mathcal{R}^\mathrm{L}$	 
\State \Call {Her}{$s,\omega,a,r^\mathrm{L},s',\tilde{\omega},\mathcal{R}^\mathrm{L}$}

\Else
\State Observe action $a$, reward $r(s,a)$, and next state $s'$ from environment
\EndIf
\State \Call{Update}{$\mathcal{R}^\mathrm{L}$,$\mathcal{Q}^{\mathrm{L}}\left( \cdot ;\theta^{\mathrm{L}}\right)$,$\mathcal{Q}^{\mathrm{L}-}\left( \cdot ;\theta^{\mathrm{L}-}\right)$}
\State $r^\mathrm{H}\gets r^\mathrm{H}+r(s,a)$
\If{$s$ is terminal state \textbf{or} $B_t=0,B_{t+1}=1$}
\If{$e < M_\epsilon$}
\State Derive $\hat{r}^\mathrm{H}$ by running policy $\pi^{\rm q}$
\Else
\State $\hat{r}^\mathrm{H}=r^\mathrm{H}$
\EndIf
\State Store transition $\left(s_0,\tilde{\omega}, \hat{r}^\mathrm{H}, s' \right) $ into $\mathcal{R}^\mathrm{H}$
\State \Call{Update}{$\mathcal{R}^\mathrm{H}$,$\mathcal{Q}^{\mathrm{H}}\left( \cdot ;\theta^{\mathrm{H}}\right)$,$\mathcal{Q}^{\mathrm{H}-}\left( \cdot ;\theta^{\mathrm{H}-}\right)$}
\If{$s$ is terminal state}
\State Delete $\left(\{-,\omega\}, -, -, \{-,\omega\} \right) $ from $\mathcal{R}^\mathrm{L}$
\State Terminate this episode
\Else
\State Update $\varOmega'$ based on \eqref{omegaspace'} 
\State $\omega\gets$ \Call{$\epsilon$-Greedy}{$s'$,$\varOmega'$,$\epsilon^\mathrm{H}$,$\mathcal{Q}^{\mathrm{H}}\left( s',\omega ;\theta^{\mathrm{H}}\right)$}  
\State $r^\mathrm{H}(s,\omega)\gets 0$, $s_0\gets s'$, $\tilde{\omega} \gets \omega$
\EndIf
\EndIf
\State $s\gets s'$
\EndFor
\State Anneal $\epsilon^{\mathrm{L}}$ and $\epsilon^{\mathrm{H}}$
\EndFor
\end{algorithmic}  
\end{algorithm}

\begin{algorithm}[ht]  
\caption{Select action $a$ or options $\omega$}
\label{alg-HDQN-greedy}
\begin{algorithmic}[1]  
\Function{$\epsilon$-Greedy}{$x$,$\mathcal{B}$,$\epsilon$,$\mathcal{Q}\left( \cdot ;\theta\right)$}
\If{random()$<\epsilon$}
\State \Return{random element from set $\mathcal{B}$}
\Else
\State \Return{$\underset{b\in \mathcal{B}}{\mathrm{arg}\max}\mathcal{Q}\left( x,b ;\theta\right) $}
\EndIf
\EndFunction
\end{algorithmic}  
\end{algorithm} 

\begin{algorithm}[ht]  
\caption{Store hindsight transitions}
\label{alg-HDQN-store}
\begin{algorithmic}[1]  
\Function{HER}{$s,\omega,a,r^\mathrm{L},s',\tilde{\omega},\mathcal{R}^\mathrm{L}$}

\State Creat hindsight transition $\left(\{s,TBD\}, a, \hat{r}^\mathrm{L}, \{s',TBD\} \right) $ where $\hat{r}^\mathrm{L}={r}^\mathrm{L}$ 
\If{$B_{t+1}=0$ \textbf{and} $\omega\neq E_{t+1}$ }
\State $\tilde{\omega}\gets E_{t+1}$, update reward $\hat{r}^\mathrm{L}$
\State Replace $TBD$ with $\tilde{\omega}$ and store hindsight transitions into $\mathcal{R}^\mathrm{L}$
\EndIf
\EndFunction
\end{algorithmic}  
\end{algorithm} 


\begin{algorithm}[ht]  
\caption{Networks updating}
\label{alg-HDQN-update}
\begin{algorithmic}[1]  
\Function{Update}{$\mathcal{R}$,$\mathcal{Q}\left( \cdot ;\theta\right)$,$\mathcal{Q}^{-}\left( \cdot ;\theta^{-}\right)$}
\State Sample a random minibatch $\mathcal{M}=\{s^{(i)},a^{(i)},r^{(i)},s'^{(i)}\}$ from $\mathcal{R}$ and set the target $$y^{\left( i \right)}=r^{\left( i \right)}+\mathcal{Q}^{-}\left( s'^{\left( i \right)},\underset{a}{\mathrm{arg}\max}Q\left(s'^{\left( i \right)},a; \theta \right) ;\theta^{-} \right) $$ 

\State Update $\mathcal{Q}$ by gradient descent on loss $L(\theta)$, where
$$		L(\theta)=\frac{1}{\left| \mathcal{M} \right|}\sum_{i\in \mathcal{M}}{\left( y^{\left( i \right)}-\mathcal{Q}\left( s^{\left( i \right)},a^{\left( i \right)} ;\theta\right) \right)}^2,$$
\EndFunction
\end{algorithmic}  
\end{algorithm}

\section{Numerical Analysis}
In order to evaluate the effectiveness of the proposed HDDQN-HER algorithm, we perform experiments based on real-world data. All the experiments are performed on a Linux server, where the DRL algorithms are implemented in Tensorflow 1.14 using Python 3.6.

\subsection{Experimental Setup}
\subsubsection{Simulated Environment}
The environment that is simulated in our experiments comprises three EBs serving the same bus route. Each of the EBs operates according to a fixed daily schedule \cite{busschedule}. Specifically, the first EB departs from the terminal station at 6:30 in the morning, and the last one departs at midnight, i.e., 0:00. An EB is scheduled to depart every $30$ minutes during this period, which means that the duration between two consecutive departures of each EB is $90$ minutes. The duration of the operating period $T_{k}^{\rm o}$ follows normal distributions with different mean values at different times of the day, i.e.,  $T_{k}^{\rm o}\sim\mathcal{N}(50,8)$ at rush hours (7:00-9:00 and 17:00-19:00) and $T_{k}^{\rm o}\sim\mathcal{N}(40,8)$ at other times.\par


The time horizon for each EB always starts with a charging period, when it arrives at the terminal station after its first departure in the morning, where a pair of consecutive charging and operating periods corresponds to one transition in the high-level SMDP. The duration of a time step is set to $10$ minutes, i.e., $\Delta t=10 \ \mathrm{min}$. The parameters of the environment are listed in Table \ref{parameters}. \par

\begin{table}[t]
\centering
\caption{Parameter configuration in the system model.}
\begin{tabular}[b]{p{2.0cm}<{\raggedright}p{1.3cm}<{\raggedright}p{4.0cm}<{\raggedright}}
\hline
\textbf{Notations}&\textbf{Values}&\textbf{Description}\\
\hline
\specialrule{0em}{1pt}{1pt}
$w_p$ & 4 &The length of the time window to look into the past prices \\
\specialrule{0em}{1pt}{1pt}
$C_{\max}$ / $D_{\max}$ & 120kW / 120kW &The maximum absolute value of charging/discharging power of EBs\\
\specialrule{0em}{1pt}{1pt}
$E_{\mathrm{min}}$ / $E_{\mathrm{max}}$    & 0kWh / 240kWh &The minimum / maximum storage constraint of the battery\\
\specialrule{0em}{1pt}{1pt}
$C^\mathrm{end}$    & 50 &The penalty for EB to enter the terminal states\\
\specialrule{0em}{1pt}{1pt}
$\kappa$ / $\kappa'$  & 0.005 / 0.0006  &The coefficients of the penalty $c^{\mathrm{tar}}(S_t,\omega_t)$/ the cost of range anxiety\\
\specialrule{0.05em}{2pt}{0pt}
\end{tabular}
\label{parameters}
\end{table}

\subsubsection{Experimental data}

\begin{figure}[t]
\centering
\includegraphics[width=0.57\linewidth]{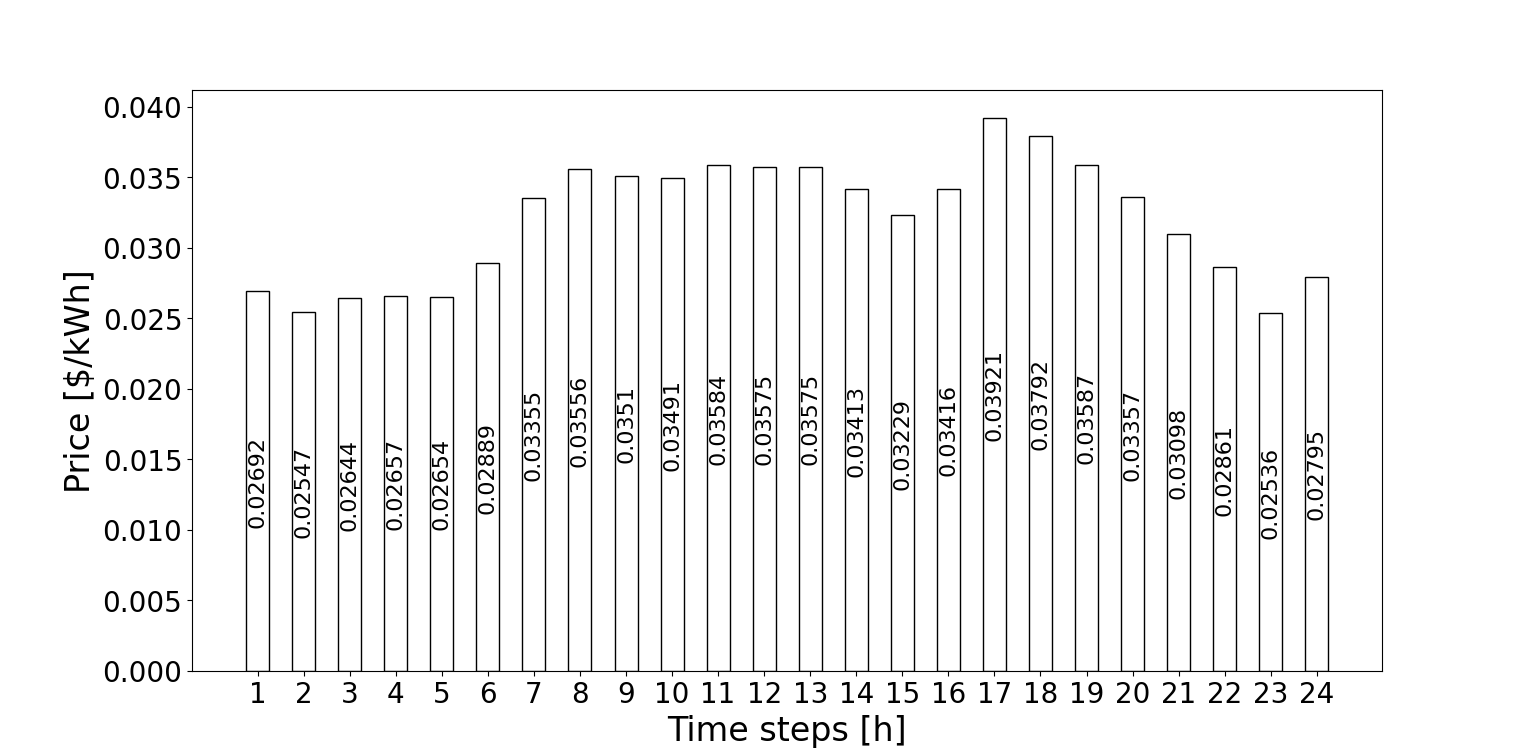}
\caption{The trajectories of the electricity prices of a typical day in the experimental data. }
\label{dataset} 
\end{figure}

We use real-world data on electricity prices to train the DRL agents. The dataset contains hourly time-varying electricity prices that reflect the wholesale market price derived from the Midcontinent Independent System Operator (MISO) delivery point \cite{MISO2023}. Since the electricity price data from one month closely approximates the statistical distribution observed over an entire year, we adopt the one-month electricity price data for training. Specifically, the data in January 2023 is used for training, and the data in the first week of February 2023 is used for testing. Fig. \ref{dataset} shows the trajectories of electricity prices on a typical day. By observing the fluctuations over time, we find that electricity prices are highest with the value of $\$0.03921$/kWh during the hours of peak demand (between 17:00 and 18:00 daily). 

\begin{figure}[t]
\centering
\includegraphics[width=0.5\linewidth]{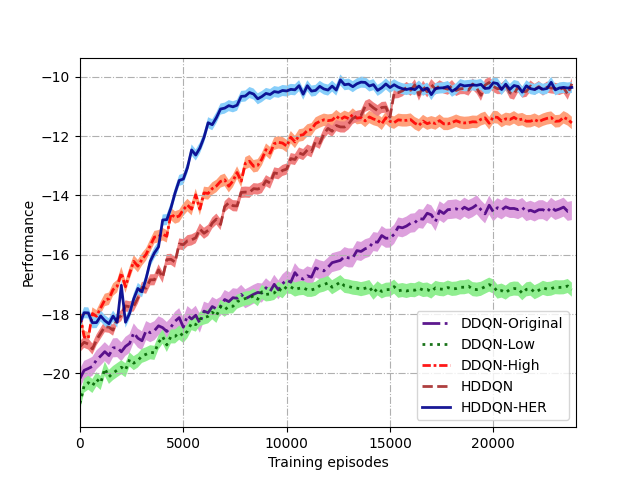}
\caption{The performance curves of the algorithms in the training process. The vertical axis corresponds to the average performance across the three EBs, and the shaded areas represent the standard errors.}
\label{training-result} 
\end{figure}


\subsubsection{Baseline algorithms}

Several baseline algorithms are simulated for performance comparison with HDDQN-HER.
\begin{enumerate}

\item \emph{DDQN-Original}: The original MDP of Section III is directly solved by DDQN to obtain a flat policy $\pi$.

\item \emph{DDQN-Low}: The low-level MDPs corresponding to the charging periods are solved by DDQN. However, the charging target is set to the maximum battery capacity, as in \cite{he2023battery,rinaldi2020mixed}, rather than being determined by the policy over options. The reward function of the low-level MDPs is similar to that defined in \eqref{rewardL}, except that $\kappa c^{\mathrm{tar}}(s,\omega)$ is replaced by $\kappa' \left(E_{\max} - E_{t_k + T_{k}^{\rm c}} \right)^2$, which measures the range anxiety. 

\item \emph{DDQN-High}: The high-level SMDP is solved by DDQN to learn the charging target. However, the low-level MDPs are solved by a fixed policy $\pi^{\rm q}$ rather than by the intra-option policy $\pi_{\omega}^*$ learned by training the low-level agent. Specifically, the charging target is realized whenever possible without considering the fluctuating electricity prices. This approach adapts the charging power at a coarse time scale, specifically once per charging period, similar to the method in \cite{9014160}. 

\item \emph{HDDQN}: Both the high-level SMDP and low-level MDPs are solved by two DDQN networks, respectively. This approach adopts the same framework as in Fig. \ref{flow}, namely the HDDQN-HER. However, the methods of overcoming non-stationarity and improving sample efficiency in Section VI.B and C are not adopted.


\end{enumerate} 

Note that a fixed low-level policy $\pi^{\rm q}$ is used in DDQN-High as well as the first learning phase in HDDQN-HER. In the experiments, we set $\pi^{\rm q}$ as follows: at each time step in a charging period, the charging power $c_{t}$ is set to the maximum value until the charging target is realized. \par

The hyper-parameters of the DDQN networks are listed in Table \ref{DRLparameters}. Note that the total number of training episodes is $24000$. The training process of HDDQN-HER is divided into two phases with the threshold of $M_{e}=6000$. Therefore, the first phase corresponds to when the training episodes obey $e<6000$, while the second phase corresponds to $6000\leq e\leq24000$. The entire training process took approximately 10 hours, while executing a single episode during the deployment process took less than 15 seconds.\par

\begin{table}[t]
\centering
\caption{Hyper-parameters of various DRL algorithms.}
\begin{tabular}[b]{p{1.6cm}<{\centering}p{1.4cm}<{\centering}p{1.2cm}<{\centering}p{1.2cm}<{\centering}p{1.6cm}<{\centering}}
\hline
\textbf{Algorithms} & \multicolumn{3}{c}{\textbf{Parameters}}\\
\specialrule{0.05em}{3pt}{3pt}
DDQN-&Q network size & Learning Rate& Batch Size & Training Episodes  \\
\specialrule{0em}{1pt}{1pt}
\hline
\specialrule{0em}{1pt}{1pt}
high-level     &256,300,100 &5e-6  &128 & 6000+18000\\
\specialrule{0em}{1pt}{1pt}
low-level      & 256,300,100 & 5e-6 & 64 & 6000+18000\\		
\specialrule{0em}{1pt}{1pt}
\hline
\specialrule{0em}{1pt}{1pt}
Original & 256,300,100 & 5e-6 & 128 & 24000\\		
\specialrule{0em}{1pt}{1pt}
High & 256,300,100 & 5e-6 & 128 & 24000\\		
\specialrule{0em}{1pt}{1pt}
Low& 400,300,100 & 1e-6 & 64 & 24000\\		
\specialrule{0em}{1pt}{1pt}
\hline
\end{tabular}
\label{DRLparameters}
\end{table}

\subsection{Experimental Results}
\subsubsection{Performance for the test set}
For each EB, we executed $100$ complete episodes based on the test set after training. We obtain the individual performance of each EB by averaging its return over the test episodes, where the return of one episode is defined in \eqref{cumulative reward}. Table \ref{result} summarizes the individual performance of each EB, as well as the average and maximum performance over the three EBs for all the algorithms. The performance rankings are consistent, where HDDQN-HER achieves the best performance, followed by HDDQN, DDQN-High, DDQN-Original, and DDQN-Low. The average performance of HDDQN-HER is better than those of the baseline algorithms by 0.097\%, 9.85\%, 28.44\%, and 39.60\%, respectively. \par


Both HDDQN and HDDQN-HER achieve much better performance than the other algorithms, demonstrating the effectiveness of our proposed framework in Fig. \ref{flow}. Note that HDDQN-HER only achieves slightly better average performance than HDDQN, while the maximum performances of the two algorithms are nearly the same. These results show that the proposed methods of overcoming non-stationarity and improving sample efficiency do not improve the final performance of HDDQN-HER much over HDDQN as long as sufficient training episodes are run. However, a significant convergence speed of the training process is observed, as it will be discussed in Section VII.B 2).\par

DDQN-Original performs worse than both HDDQN-HER and HDDQN, since it struggles to learn an optimal policy by directly solving the original MDP for a long horizon. As the penalty $C^{\mathrm{end}}$ for entering the terminal states is sparse, a long sequence of highly-specific actions must be executed prior to observing a terminal reward, which makes the learning task challenging. Meanwhile, the hierarchical framework in HDDQN-HER provides an intrinsic reward for the low-level MDPs through options, which substantially improves the sample efficiency. \par 

HDDQN-HER outperforms DDQN-Low due to the more reasonable charging target in the reward function. The charging target of DDQN-Low is set to the maximum battery capacity, while those of HDDQN-HER are learned by solving the high-level SMDP. Therefore, HDDQN-HER can adapt to the electricity price changes throughout the day and consider the long-range effects of decisions. In addition, compared to HDDQN-HER, DDQN-High discards the low-level DDQN network, which prevents it from adapting EB charging power at a fine time scale to fluctuating electricity prices. Consequently, it cannot learn optimal policies for the low-level MDPs. Finally, comparisons with both DDQN-High and DDQN-Low demonstrate the importance of both levels of DDQN networks for the proposed algorithm. Omitting either level would hinder the algorithm's ability to achieve optimal performance. 

\begin{table}[t]
\small
\centering
\caption{The individual, average, and maximum performance of the algorithms across three EBs.}
\label{result}
\begin{tabular}{|l|ccccc|}
\hline
\multirow{2}{*}{Algorithms} & \multicolumn{5}{c|}{Performance}                                                                                                                              \\ \cline{2-6} 
& \multicolumn{1}{c|}{EB A}    & \multicolumn{1}{c|}{EB B} & \multicolumn{1}{c|}{EB C} & \multicolumn{1}{c|}{Max} & \multicolumn{1}{c|}{Average}  \\ \hline
HDDQN-HER                       & \multicolumn{1}{c|}{-10.23}  & \multicolumn{1}{c|}{-10.48}     & \multicolumn{1}{c|}{-10.31}     & \multicolumn{1}{c|}{-10.23}   & \multicolumn{1}{c|}{-10.34}                \\ \hline
HDDQN                       & \multicolumn{1}{c|}{-10.23}  & \multicolumn{1}{c|}{-10.50}     & \multicolumn{1}{c|}{-10.32}     & \multicolumn{1}{c|}{-10.23}   & \multicolumn{1}{c|}{-10.35}                \\ \hline
DDQN-High                    & \multicolumn{1}{c|}{-11.25}  & \multicolumn{1}{c|}{-11.67}     & \multicolumn{1}{c|}{-11.48}     & \multicolumn{1}{c|}{-11.25}   & \multicolumn{1}{c|}{-11.47}                \\ \hline
DDQN-Original        & \multicolumn{1}{c|}{-14.08} & \multicolumn{1}{c|}{-14.67}     & \multicolumn{1}{c|}{-14.59}     & \multicolumn{1}{c|}{-14.08}   & \multicolumn{1}{c|}{-14.45}           \\ \hline
DDQN-Low             & \multicolumn{1}{c|}{-16.92} & \multicolumn{1}{c|}{-17.38}     & \multicolumn{1}{c|}{-17.05}     & \multicolumn{1}{c|}{-16.92}   & \multicolumn{1}{c|}{-17.12}         \\ \hline
\end{tabular}
\end{table}

\subsubsection{Convergence Properties}
Fig. \ref{training-result} shows the performance curves of the algorithms, which are obtained by periodically evaluating the policies during training. Specifically, we ran $10$ test episodes for every $100$ training episode, where the X-axis is the number of training episodes and the Y-axis is the performance corresponding to the average return over the latest $10$ test episodes. The shaded areas indicate the standard errors across the three EBs.  

The proposed HDDQN-HER converges first at approximately $8500$ episodes, followed by DDQN-Low at approximately $10000$ episodes due to its simple framework and short time horizon. However, the performance of DDQN-Low after convergence is inferior to those of the other algorithms. DDQN-High also converges relatively fast at approximately $12000$ episodes. By contrast, without the methods of overcoming non-stationarity and improving sample efficiency in HDDQN-HER, HDDQN converges slowly at around $15000$ episodes. Finally, the convergence of DDQN-Original is the slowest, which converges at around $18000$ episodes. This result further demonstrates the low sample and learning efficiency, when directly solving a long-range planning problem by RL. The shaded areas of all the algorithms are relatively small, which shows that all the algorithms perform consistently for different EBs.\par



\subsubsection{Charging Schedule Results}
\begin{figure*}[t]
\centering
\subfigure[charging scheduling of HDDQN-HER]{
\label{chargingHDQN} 
\includegraphics[height=3.6cm,width=0.45\textwidth]{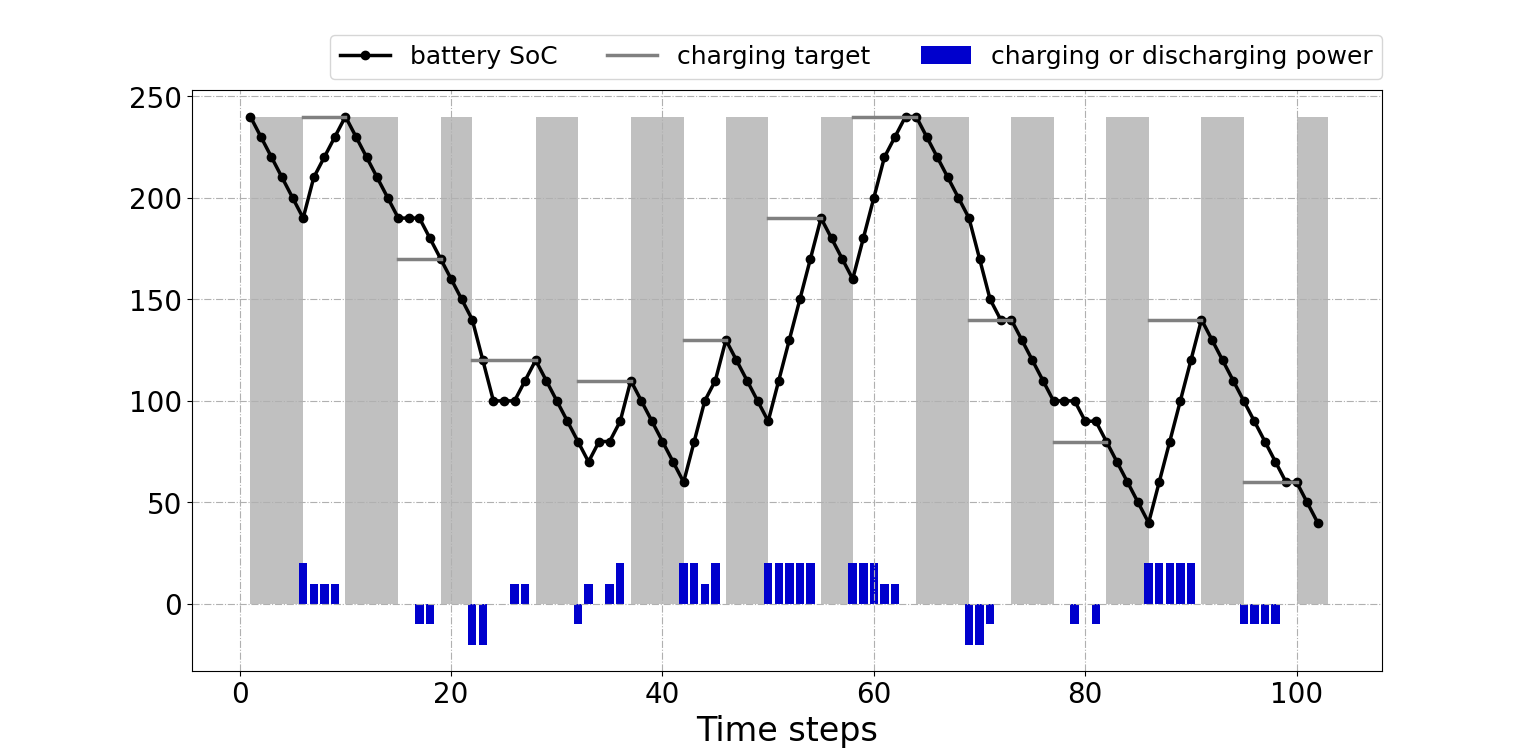}}
\subfigure[costs of HDDQN-HER]{
\label{costHDQN} 
\includegraphics[height=3.6cm,width=0.45\textwidth]{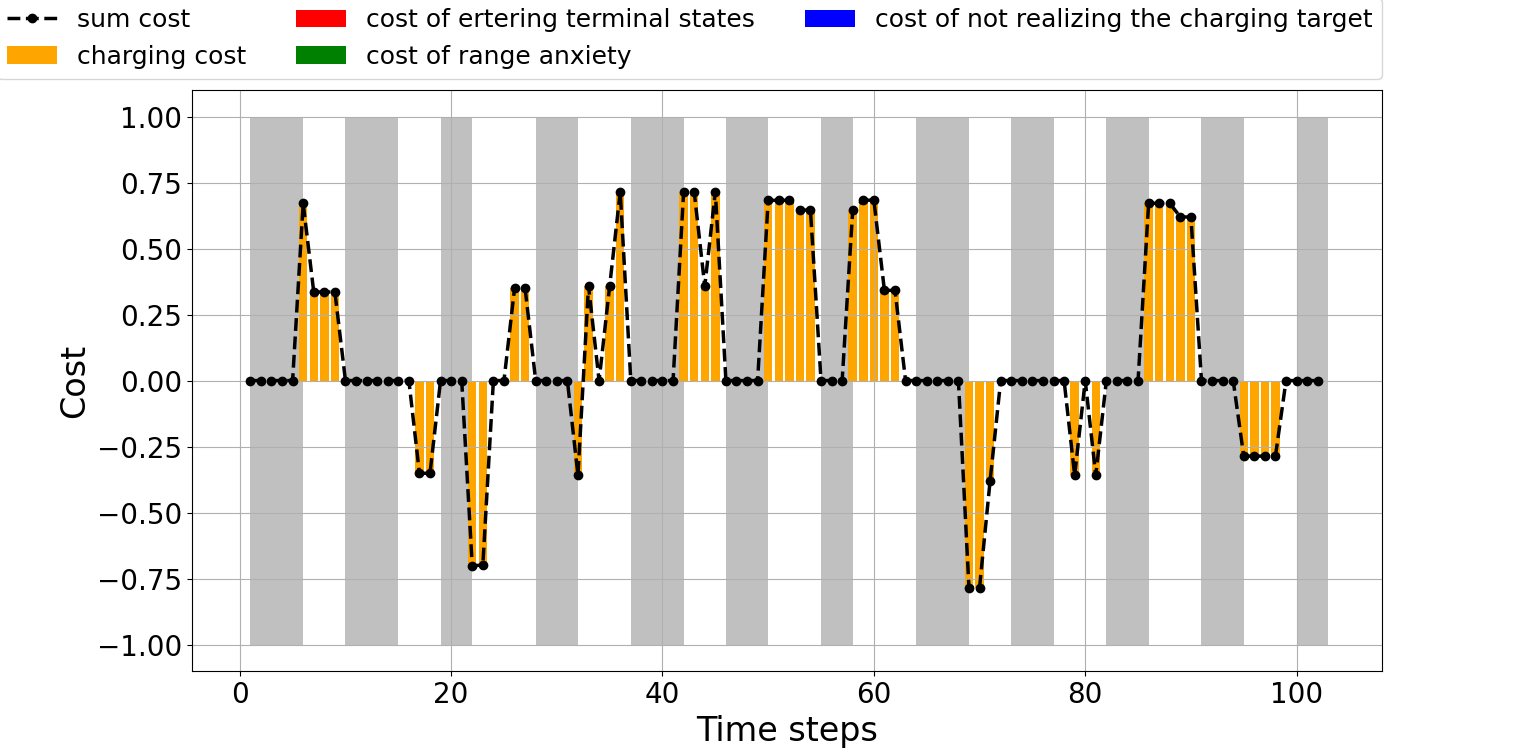}}
\subfigure[charging scheduling of DDQN-Original]{
\label{chargingDQNori} 
\includegraphics[height=3.6cm,width=0.45\textwidth]{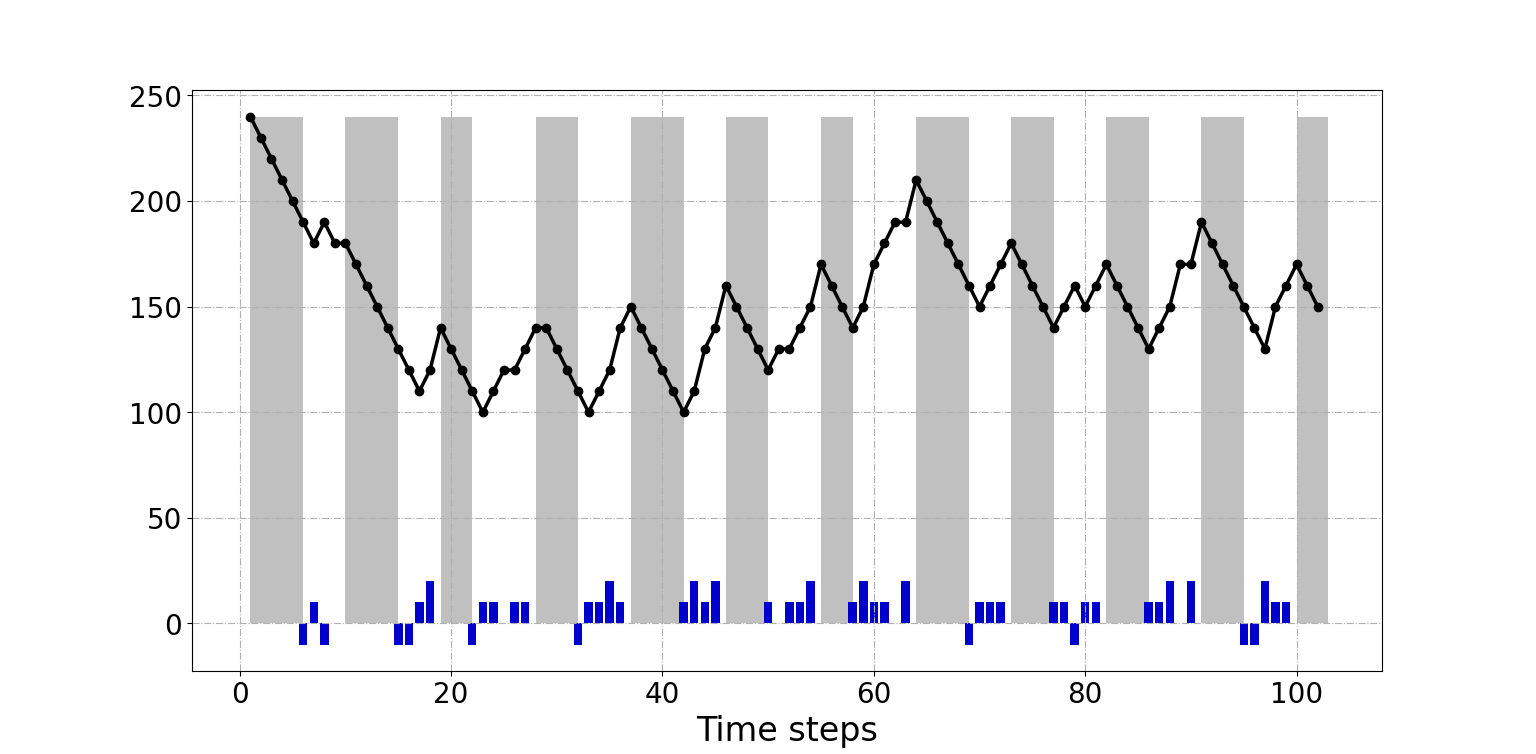}}
\subfigure[costs of DDQN-Original]{
\label{costDQNori} 
\includegraphics[height=3.6cm,width=0.45\textwidth]{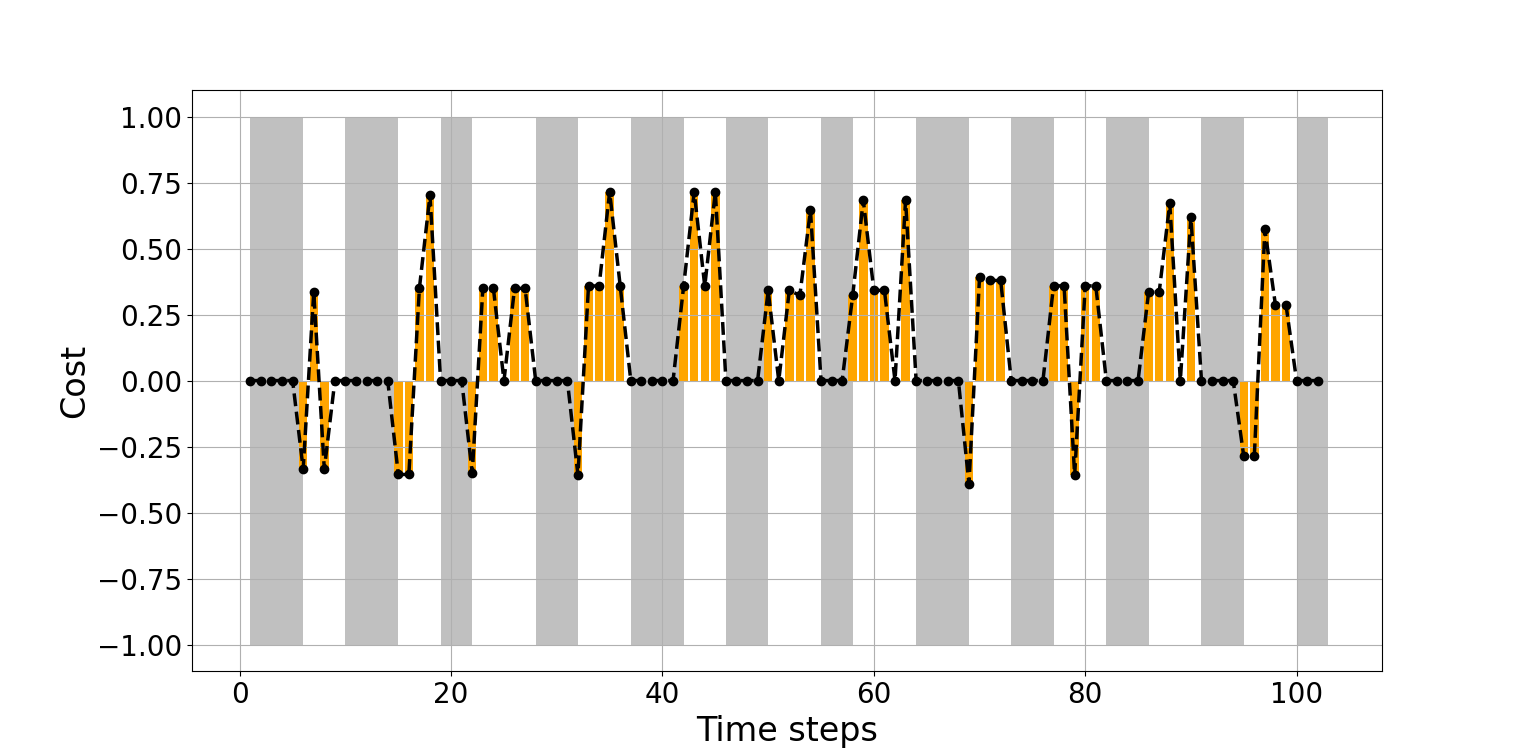}}
\subfigure[charging scheduling of DDQN-Low]{
\label{chargingDQNlow} 
\includegraphics[height=3.6cm,width=0.45\textwidth]{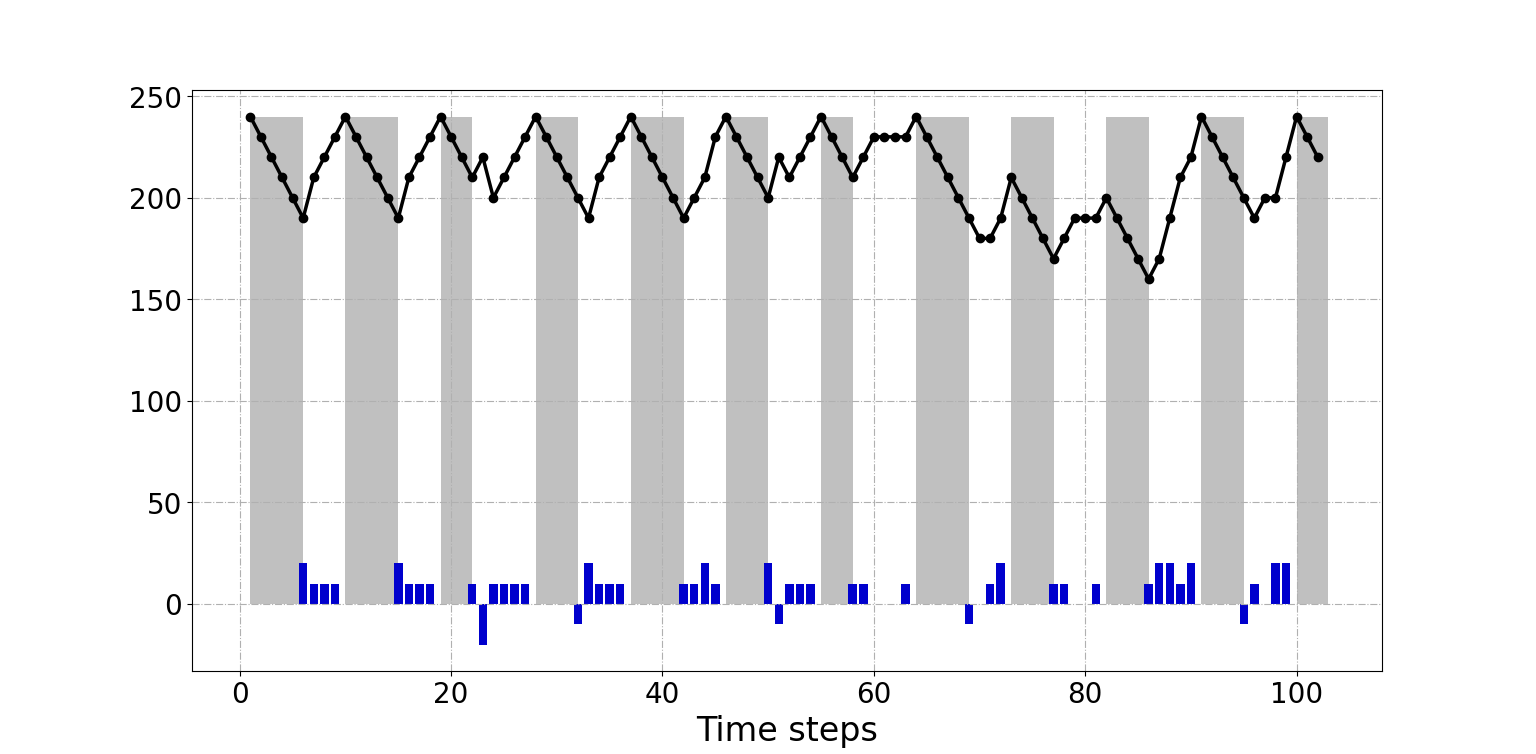}}
\subfigure[costs of DDQN-Low]{
\label{costDQNlow} 
\includegraphics[height=3.6cm,width=0.45\textwidth]{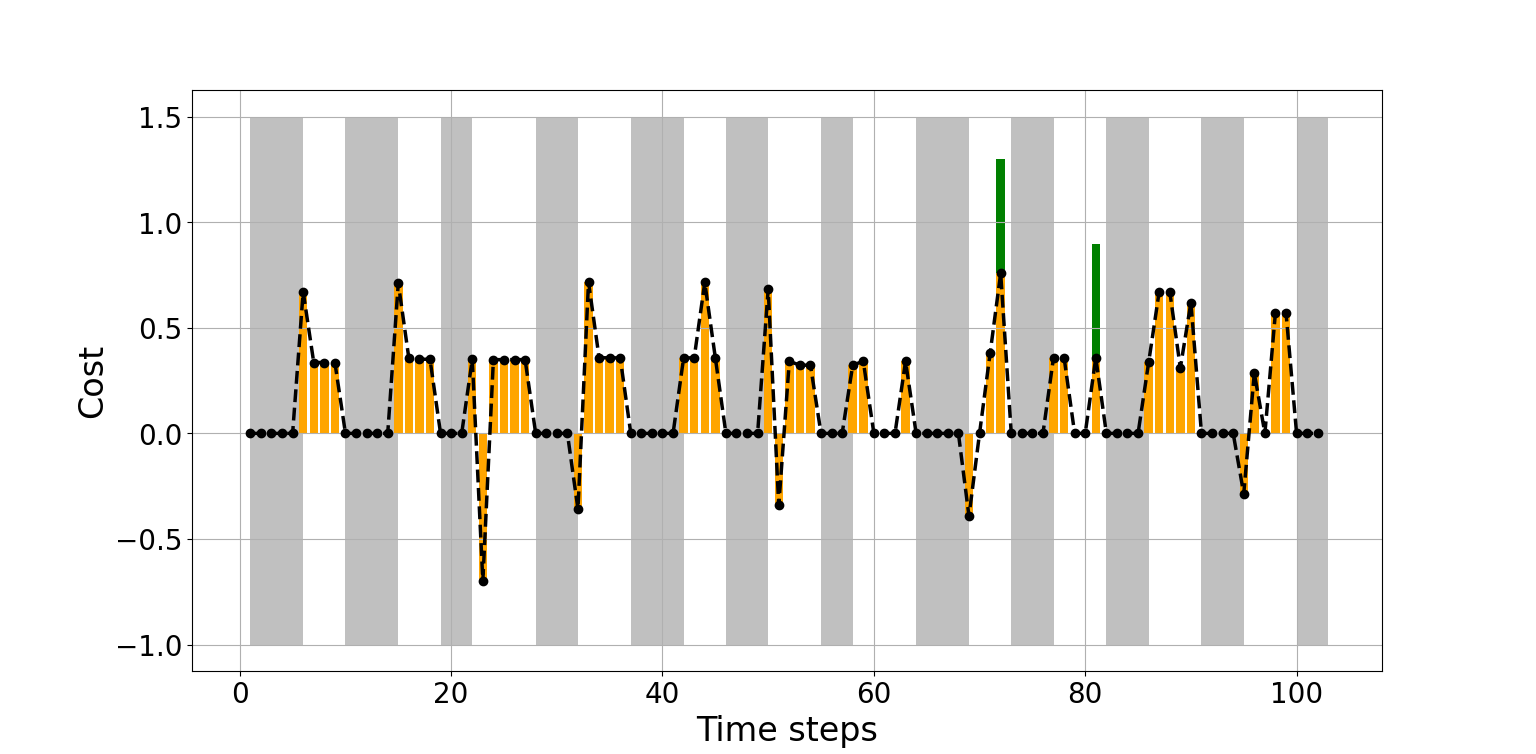}}

\caption{A specific test episode on an EB.}
\label{trajectory} 
\end{figure*}

Fig. \ref{trajectory} illustrates the detailed charging schedule decisions and corresponding costs for EB A at each time step in a typical episode based on the test set. The grey and white regions indicate the operating and charging periods, respectively. Fig. \ref{chargingHDQN}, \ref{chargingDQNori}, and \ref{chargingDQNlow} show the charging power at each time step and the EB's battery trajectory. Notice that there is a horizontal line for each charging period in Fig. \ref{chargingHDQN}, which represents the learned charging target. In Fig. \ref{costHDQN}, \ref{costDQNori}, and \ref{costDQNlow}, the sum cost trajectory is displayed as a curve, where the sum cost at time step $t$ is derived by the negative reward $-r(S_t, A_t)$ in \eqref{c}. Since the sum cost consists of two components, i.e., the charging cost and the cost of entering terminal states, we use the bar chart to visualize the individual costs in detail. Moreover, the cost components in the reward function of low-level MDPs are displayed, which include the cost of not realizing the charging target $\kappa c^{\mathrm{tar}}(S_t,\omega_t)$ in HDDQN-HER, and the cost of range anxiety $\kappa' \left(E_{\max} - E_{t_k + T_{k}^{\rm c}} \right)^2$ in DDQN-Low. \par 



Fig. \ref{costHDQN}, \ref{costDQNori}, and \ref{costDQNlow} show that there is no cost for entering the terminal states, which means that the basic requirement of ensuring sufficient battery energy can be met by all the algorithms. A closer examination of Fig. \ref{chargingHDQN} and \ref{chargingDQNori} shows that the battery trajectory curve of HDDQN-HER fluctuates much more than that of DDQN-Original, which is because the high-level DDQN in the proposed algorithm can better learn the fluctuating trends of electricity prices in a full day. Therefore, the high-level agent decides to charge more when the electricity price is low in the afternoon from 14:00 to 16:00, and sell more electricity when the price is the highest, namely from 17:00 to 19:00 to minimize its charging cost. Meanwhile, DDQN-Original is unable to learn such an efficient policy, which leads to the relatively high charging cost seen in Fig. \ref{costDQNori}. Additionally, Fig. \ref{chargingHDQN} shows that HDDQN-HER can realize the charging target in every charging period, and thus no corresponding cost is incurred, as shown in Fig. \ref{costHDQN}. Finally, as shown in Fig. \ref{chargingDQNlow}, the battery SoC curve of DDQN-Low continues to remain high, since the charging target is the constant full battery capacity. Due to the cost of range anxiety, the agent tends to fully charge the EB in most charging periods, which is an essential reason why DDQN-Low has the highest charging cost. It can be observed in Fig. \ref{costDQNlow} that the cost of range anxiety is always zero except in the $8$-th and $9$-th charging periods, which means that the EB is always fully charged except in these two periods.

\section{Conclusions}
We have studied how to use HDRL techniques to learn optimal charging schedules for EBs in a system model giving full cognizance to both the charging and operating periods. An MDP model has been conceived, which has been reformulated into a two-level decision process in order to overcome the challenges of long-range multi-phase planning with sparse rewards. It has been proved that the flat policy created by superimposing the optimal high-level and low-level policies performs as well as the optimal policy of the original MDP. The HDDQN-HER algorithm has been conceived for training the two levels of agents simultaneously, in which compelling methods leveraging the idea of HER are proposed to deal with the non-stationarity encountered in training the high-level agent and to improve the sample efficiency in training the low-level agent. Finally, experiments have been performed to demonstrate that the proposed algorithm achieves better performance than the baseline algorithms. 

In our future work, the HDRL framework will be extended to consider charging schedules for EB fleets, where the high-level bus scheduling decisions that are made at a coarser time-scale can be optimized jointly with the low-level charging power decisions made at a finer time-scale. Our ultimate goal is to leash all solutions of the multi-component Pareto-front having a gradually increased number of parameters in the objective function. Additionally, we will integrate safe RL techniques to provide formal guarantees that the agent does not enter terminal states. This enhancement will further improve the robustness of our approach in real-world deployments. Finally, we plan to incorporate real-world EB operational data instead of simulated data in our experiments to better capture real-world uncertainties and refine policy performance under practical conditions.\par

\begin{appendices}

\section{Proof for Theorem \ref{thm_1}}
\begin{proof}
\begin{itemize}
    \item Reparation stage: Without loss of generality, the R.H.S. of \eqref{thm} in Theorem \ref{thm_1} can be derived as
\end{itemize}

\begin{align}
\label{proof 0}
&V_{\pi ^*}(s)\stackrel{({\rm a})}{=}\underset{\pi}{\max}V_{\pi}(s)
\stackrel{({\rm b})}{=}\underset{\pi}{\max}\mathbb{E} _{\pi}\left[ \sum_{t=0}^{T-1}{r\left( S_{t},A_{t} \right) }|S_0=s \right] \IEEEnonumber \\ 
&\stackrel{({\rm c})}{=}\underset{\pi}{\max}\mathbb{E} _{\pi}\left[ \sum_{k=0}^{K-1}{\sum_{l=0}^{{T}_{k}^{\rm a}-1}{r\left( S_{t_{k}+l},A_{t_{k}+l} \right)}}|S_0=s \right] ,\forall s\in \mathcal{S} ,
\end{align}

\noindent where (a) is because $\pi^*$ is the optimal flat policy; (b) is derived by definition of $V_{\pi}(s)$; (c) reformulates the reward summation by grouping rewards according to periods.

When a charging period $k\in\{0,1,\cdots,K-1\}$ terminates after a random number of $T_{k}^{\rm c}$ time steps, we denote the SoC of the battery when following the optimal flat policy $\pi^{*}$ as $E_{t_{k}+T_{k}^{\rm c}}^{*}$. \par 

In the following, we will prove Theorem 1 in \textbf{two steps}. Specifically, we decouple $\pi^*$ into a high-level policy over options $\hat{\mu}$ and a low-level intra-option policy $\hat{\pi}_\omega$. Thus, the SoC levels $E_{t_{k}+T_{k}^{\rm c}}^{*}$,  $\forall k\in\{0,1,\cdots,K-1\}$ can be regarded as the charging targets prescribed by $\hat{\mu}$ , i.e., 
\begin{equation}
\label{hat mu target}
\hat{\mu}(S_{t_{k}})=\hat{\omega}=E_{t_{k}+T_k^{\rm c}}^{*}, \forall k\in\{0,1,\cdots,K-1\},
\end{equation}
\noindent while the charging target $\hat{\omega}$ is fulfilled by the policy $\hat{\pi}_{\hat{\omega}}$, i.e.,
\begin{equation}
\label{hat pi target}
\hat{\pi}_{\hat{\omega}}(S_{t})=\pi^{*}(S_{t}), \forall t\in\{t_k,\cdots,t_k+T_k^{\rm c}-1\}.
\end{equation}

\begin{itemize}
    \item In the first step, we will prove that the optimal low-level intra-option policy $\pi_{\hat{\omega}}^*$ achieves the same performance as $\hat{\pi}_{\hat{\omega}}$ does when their charging targets are both prescribed by $\hat{\mu}$ in \eqref{hat mu target}, i.e.,
\end{itemize}

\begin{align}
\label{prooflow-level}
\mathbb{E}_{{\pi}^*_{\hat{\omega}}}\left[\sum_{l=0}^{{T}_{k}^{\rm c}}{r\left( S_{t_{k}+l},A_{t_{k}+l} \right)}|S_{t_{k}}=s  \right] \IEEEnonumber \\
=\mathbb{E}_{\hat{\pi}_{\hat{\omega}}}\left[\sum_{l=0}^{{T}_{k}^{\rm c}}{r\left( S_{t_{k}+l},A_{t_{k}+l} \right)}|S_{t_{k}}=s  \right].
\end{align} 
For this purpose, we adopt the proof by contradiction method and assume that \eqref{prooflow-level} does not hold. Firstly, we have

\begin{align}
\label{proof VL}
&\mathbb{E}_{{\pi}^*_{\hat{\omega}}}\left[\sum_{l=0}^{{T}_{k}^{\rm c}}{r\left( S_{t_{k}+l},A_{t_{k}+l} \right)}|S_{t_{k}}=s  \right] \IEEEnonumber \\
&\stackrel{({\rm a})}{=}\mathbb{E}_{{\pi}^*_{\hat{\omega}}}\left[\sum_{l=0}^{{T}_{k}^{\rm c}}{r^{\rm L}\left( S_{t_{k}+l},\hat{\omega},A_{t_{k}+l} \right)}|S_{t_{k}}=s  \right] \IEEEnonumber \\
&\stackrel{({\rm b})}{=}\underset{{\pi} _{\hat{\omega}}}{\max}\mathbb{E} _{\pi _{\hat{\omega}}}\left[ \sum_{l=0}^{{T}_k^{\rm c}}{r^{\mathrm{L}}\left( S_{t_k+l},\hat{\omega},A_{t_k+l} \right)} |S_{t_{k}}=s  \right] \IEEEnonumber \\
& \stackrel{({\rm c})}{\geqslant}\mathbb{E} _{\hat{\pi}_{\hat{\omega}}}\left[ \sum_{l=0}^{{T}_k^{\rm c}}{r^{\rm L}\left( S_{t_k+l},\hat{\omega},A_{t_k+l} \right)} |S_{t_{k}}=s \right] \IEEEnonumber \\
& \stackrel{({\rm d})}{=}\mathbb{E} _{\hat{\pi}_{\hat{\omega}}}\left[ \sum_{l=0}^{{T}_k^{\rm c}}{r\left( S_{t_k+l},A_{t_k+l} \right)} |S_{t_{k}}=s \right], 
\end{align}
\noindent where (a) and (d) are valid, because there is no penalty for failing to realize the charging target in \eqref{rewardL}, i.e., $c^{\mathrm{tar}}(s,\omega)=0$ since both $\pi_{\hat{\omega}}^*$ and $\hat{\pi}_{\hat{\omega}}$ can realize the charging target $\hat{\omega}$; (b) and (c) are true according to the definition of the optimal value function in \eqref{optimal low value}.\par

Since we assume that \eqref{prooflow-level} does not hold, (c) in \eqref{proof VL} is a strict inequality due to the property of the max operator, which means that we have

\begin{align}
\label{contradict_step1}
& \mathbb{E} _{\hat{\mu}}\mathbb{E} _{\pi_{\hat{\omega}}^*}\left[\sum_{k=0}^{K-1} \sum_{l=0}^{{T}_k^{\rm c}}{r\left( S_{t_k+l},A_{t_k+l} \right)} |S_{0}=s \right]  \IEEEnonumber \\
& \stackrel{({\rm a})}{>}\mathbb{E} _{\hat{\mu}}\mathbb{E} _{\hat{\pi}_{\hat{\omega}}}\left[\sum_{k=0}^{K-1} \sum_{l=0}^{{T}_k^{\rm c}}{r\left( S_{t_k+l},A_{t_k+l} \right)} |S_{0}=s \right] \IEEEnonumber \\
&\stackrel{({\rm b})}{=}\mathbb{E} _{\pi^{*}}\left[\sum_{k=0}^{K-1} \sum_{l=0}^{{T}_k^{\rm c}}{r\left( S_{t_k+l},A_{t_k+l} \right)} |S_{0}=s \right] \IEEEnonumber\\
&\stackrel{({\rm c})}{=}\mathbb{E} _{\pi^{*}}\left[\sum_{k=0}^{K-1} \sum_{l=0}^{{T}_k^{\rm a}-1}{r\left( S_{t_k+l},A_{t_k+l} \right)} |S_{0}=s \right], 
\end{align} 
\noindent where (a) is derived from \eqref{proof VL} with strict inequality; (b) is due to the fact that $\pi^*$ is the superimposition of $\hat{\mu}$ and $\hat{\pi}_{\hat{\omega}}$; (c) is because of the zero rewards during the operating periods from ${T}_k^{\rm c}$ to ${T}_k^{\rm a}-1$, since there is no charging cost in the operating period. Moreover, the optimal flat policy $\pi^*$ ensures that there is no penalty for entering the terminal states, i.e., $C^{\mathrm{end}}=0$  in \eqref{c}. \par

Furthermore, \eqref{contradict_step1} implies that the flat policy created by superimposing the policy $\hat{\mu}$ and the optimal intra-option policy $\pi_{\hat{\omega}}^*$ performs better than the optimal policy $\pi^{*}$ for the original MDP, which cannot be true by the definition of $\pi^{*}$. Therefore, \eqref{contradict_step1} contradicts to \eqref{proof 0}, and we have thus proved that \eqref{prooflow-level} holds.

\begin{itemize}
    \item In the second step, we will prove that the optimal high-level policy over options $\mu^*$ achieves the same performance as $\hat{\mu}$ does when the low-level intra-option policy is $\pi_{\omega}^*$, i.e.,
\end{itemize}

\begin{align}
\label{proof_2}
&\mathbb{E} _{\mu^*}\mathbb{E} _{\pi_{\omega^*}^*}\left[\sum_{k=0}^{K-1} r^{\mathrm{H}}\left( S_{t_k},\omega^* \right) |S_{0}=s \right]  \IEEEnonumber \\
& =\mathbb{E} _{\hat{\mu}}\mathbb{E} _{\pi_{\hat{\omega}}^*}\left[\sum_{k=0}^{K-1} r^{\mathrm{H}}\left( S_{t_k},\hat{\omega} \right)  |S_{0}=s \right]  .
\end{align}

In order to prove \eqref{proof_2}, we also adopt the proof by contradiction method and assume that \eqref{proof_2} does not hold. Thus, we have
\begin{align}
\label{proof VH}
&\mathbb{E} _{\mu^*}\mathbb{E} _{\pi_{\omega^*}^*}\left[\sum_{t=0}^{T-1} {r\left( S_{t},A_{t}\right) } |S_{0}=s \right]  \IEEEnonumber \\
& \stackrel{({\rm a})}{=}\mathbb{E} _{\mu _{}^*}\mathbb{E} _{\pi _{\omega^*}^*}\left[ \sum_{k=0}^{K-1}{r^{\mathrm{H}}\left( S_{t_k},\omega^* \right)} |S_{{0}}=s  \right] \IEEEnonumber \\
& \stackrel{({\rm b})}{=}\underset{\mu}{\max}\mathbb{E} _{\mu}\mathbb{E} _{\pi^{*} _{\omega}}\left[ \sum_{k=0}^{K-1}{r^{\rm H}\left( S_{t_k},\omega \right)} |S_{{0}}=s  \right] \IEEEnonumber \\
& \stackrel{({\rm c})}{>}\mathbb{E} _{\hat{\mu}}\mathbb{E} _{\pi^* _{\hat{\omega}}}\left[ \sum_{k=0}^{K-1}{r^{\rm H}\left( S_{t_k},\hat{\omega} \right)} |S_{{0}}=s \right] \IEEEnonumber \\
& \stackrel{({\rm d})}{=}\mathbb{E} _{\hat{\mu}}\mathbb{E} _{\hat{\pi} _{\hat{\omega}}}\left[ \sum_{k=0}^{K-1}{r^{\rm H}\left( S_{t_k},\hat{\omega} \right)} |S_{{0}}=s \right]  \IEEEnonumber \\
&\stackrel{({\rm e})}{=}\mathbb{E} _{{\pi^*}}\left[\sum_{t=0}^{T-1} {r\left( S_{t},A_{t}\right) } |S_{0}=s \right]\stackrel{({\rm f})}{=}V_{\pi^*}(s), 
\end{align}
\noindent where (a) is derived by definition of the high-level reward in \eqref{optionreward}; (b) is derived by definition of the optimal value function in \eqref{value_high 2}; (c) is a strict inequality due to the property of the max operator and the assumption that \eqref{proof_2} does not hold; (d) is due to \eqref{prooflow-level}; (e) is because the policies $\hat{\mu}$ and $\hat{\pi}_{\hat{\omega}}$ superimpose the optimal flat policy $\pi^*$; (f) is by definition of the value function in \eqref{value}. \par

Finally, \eqref{proof VH} implies that the flat policy created by superimposing ${\mu^*}$ and $\pi_{{\omega}^*}^*$ performs better than the optimal policy $\pi^{*}$ for the original MDP, which cannot be true by the definition of $\pi^{*}$. Therefore, \eqref{proof VH} contradicts to \eqref{proof 0}, and we have thus proved that \eqref{proof_2} holds. Moreover, if \eqref{proof_2} holds, (c) in \eqref{proof VH} should be an equality. Notably, the first term in \eqref{proof VH} corresponds to the right-hand side of \eqref{thm}, while the last term corresponds to its left-hand side, which means that \eqref{thm} in Theorem \ref{thm_1} holds. \par

\end{proof}

\end{appendices}

\bibliography{author}
\bibliographystyle{IEEEtran}

\end{document}